\documentclass{article}

\PassOptionsToPackage{numbers, compress}{natbib}
\usepackage[preprint]{neurips_2026}

\makeatletter
\renewcommand{\@noticestring}{Accepted at the 40th Conference on Neural Information Processing Systems (NeurIPS 2026).}
\makeatother

\usepackage[utf8]{inputenc}
\usepackage[T1]{fontenc}
\usepackage{hyperref}
\usepackage{url}
\usepackage{booktabs}
\usepackage{amsfonts}
\usepackage{nicefrac}
\usepackage{microtype}
\usepackage{xcolor}

\usepackage{amsmath,amssymb}
\usepackage{graphicx}
\usepackage{multirow}
\usepackage{subcaption}
\usepackage{enumitem}
\usepackage{algorithm}
\usepackage{algorithmic}
\usepackage{cleveref}
\usepackage{pifont}
\usepackage{rotating}
\usepackage{titletoc}

\hypersetup{
	colorlinks=true,
	linkcolor=red!60!black,
	citecolor=green!40!black,
	urlcolor=blue
}
\newcommand{\cmark}{\textcolor{teal}{\ding{51}}}
\newcommand{\xmark}{\textcolor{red}{\ding{55}}}

\graphicspath{{./}}

\title{BAT3R: Robust Online 3D Reconstruction with Bayesian Adaptive State Updates}

\author{%
  Lijian Li$^{1}$\thanks{Equal contribution.}\quad
  Yuanpeng He$^{2,3,*}$\thanks{Corresponding authors: Yuanpeng He
  (\texttt{heyuanpeng@stu.pku.edu.cn}) and Chi-Man Pun
  (\texttt{cmpun@um.edu.mo}).}\quad
  Yudian Zheng$^{1}$\\[0.5ex]
  \textbf{Chung-ju Huang}$^{2}$\quad
  \textbf{Linyu Li}$^{2,3}$\quad
  \textbf{Dongming Jin}$^{2,3}$\\[0.5ex]
  \textbf{Wenpin Jiao}$^{2,3}$\quad
  \textbf{Zhi Jin}$^{2,3}$\quad
  \textbf{Chi-Man Pun}$^{1,\dagger}$\\[1ex]
  \normalfont\small
  $^{1}$University of Macau, Macau SAR, China\\
  $^{2}$School of Computer Science, Peking University, Beijing, China\\
  $^{3}$Key Laboratory of High Confidence Software Technologies (Peking University),\\
  \normalfont\small Ministry of Education, Beijing, China
}
\hypersetup{%
  pdftitle={BAT3R: Robust Online 3D Reconstruction with Bayesian Adaptive State Updates},
  pdfauthor={Lijian Li, Yuanpeng He, Yudian Zheng, Chung-ju Huang, Linyu Li, Dongming Jin, Wenpin Jiao, Zhi Jin, Chi-Man Pun}
}

\begin{document}

\maketitle
\begin{abstract}
Feed-forward transformer models have driven remarkable progress in 3D vision, yet their quadratic complexity renders them impractical for long video sequences. Streaming 3D reconstruction addresses this bottleneck by processing frames sequentially with constant memory. However, existing recurrent architectures inevitably suffer from progressive degradation over long sequences, as they fail to account for the intrinsic reliability of observations and the accumulated confidence of the historical state, leading to severe error accumulation. To address these issues, we introduce BAT3R, a training-free Bayesian adaptation framework comprising two synergistic modules: State Innovation Gating (SIG) and Bayesian-guided Adaptive State Updating (BASU). SIG acts as a distributional regularizer that anchors state evolution to its geometric initialization, effectively arresting representational drift during long-term inference. Building on this, BASU reformulates state update as a recursive Bayesian inference process within a Kalman filtering framework. By jointly modeling the temporal reliability of observations via information-theoretic gain and spatial relevance via cross-attention maps, BASU achieves an asymmetric spatio-temporal state fusion: well-consolidated historical regions are protected from redundant updates, while genuinely novel scene content is aggressively incorporated. Extensive experiments on standard benchmarks (7-Scenes, TUM, ScanNet, KITTI, etc.) demonstrate that BAT3R consistently outperforms state-of-the-art online 3D reconstruction methods, with the most significant gains observed on long sequences and complex scenes.
\end{abstract}

\section{Introduction}

\begin{figure*}[htbp]
	\centering
	\includegraphics[width=\textwidth]{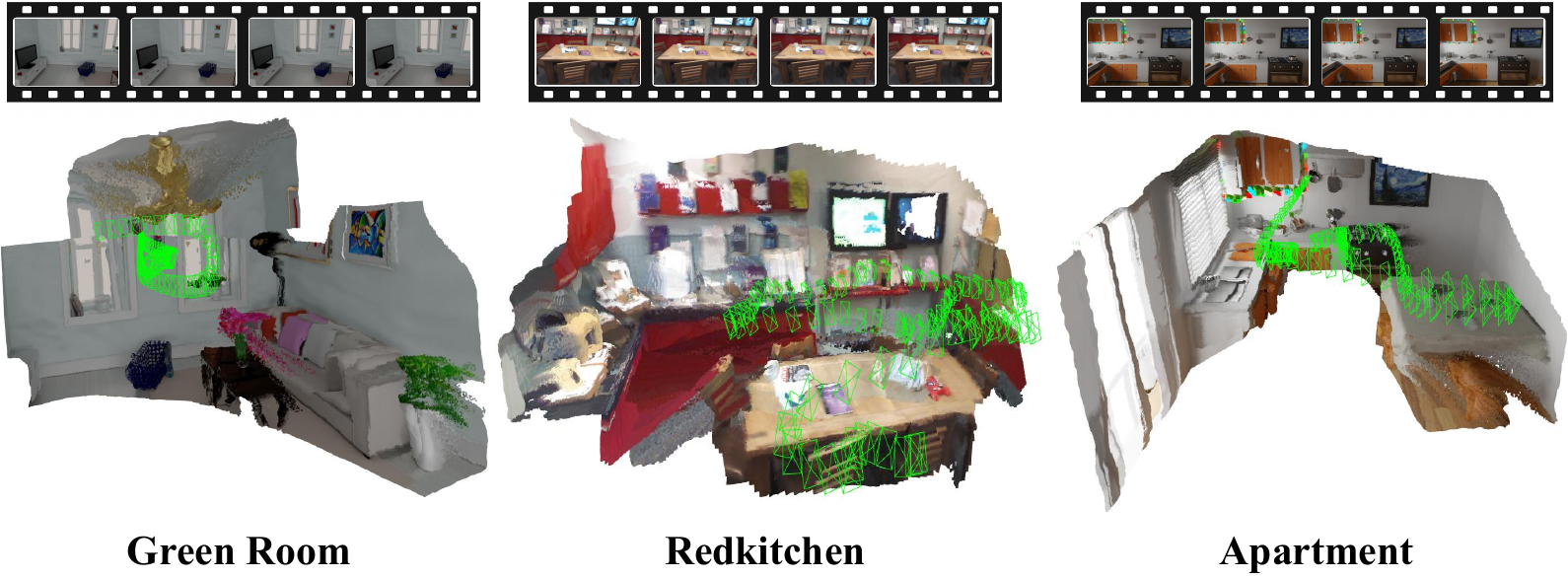}
	\caption{\textbf{Robust Online 3D Reconstruction.} Given a stream of RGB images as input, our \textbf{BAT3R} framework enables dense 3D reconstruction in an online, continuous manner, estimating both camera parameters and dense 3D geometry with each incoming frame.}
	\label{fig:prefix}
\end{figure*}

Real-time 3D scene understanding from monocular video streams, as illustrated in Fig. \ref{fig:prefix}, is a foundational challenge in computer vision, with direct applications in robotic navigation and augmented reality . Recent feed-forward transformer models \cite{wang2024dust3r, wang2025vggt, zhang2025advances}, such as DUSt3R \cite{wang2024dust3r} and VGGT \cite{wang2025vggt}, have achieved remarkable reconstruction accuracy through dense global attention \cite{torresani2008nonrigid} that jointly reasons across all input frames; however, the computational and memory cost of these methods grows quadratically with sequence length, rendering them impractical for continuous perception over arbitrarily long videos. Streaming reconstruction methods \cite{chen2025human3r, chen2025long3r, xiang2025gaussianroom, yuan2025test3r} address this scalability issue by incrementally processing frames and compressing historical context into a fixed-length recurrent latent state, thereby achieving constant-memory spatial perception; yet existing recurrent frameworks \cite{cabon2025must3r, chen2025long3r, wang20253d} inevitably suffer from long-range performance decay, as the continuous update of scene states without revisiting prior observations causes cumulative structural drift and progressive distortion of global geometry.

Test-Time Training (TTT) has recently emerged as a compelling paradigm for efficient, linear-complexity sequence modeling. In the context of spatial perception, TTT3R \cite{chen2025ttt3r} pioneers this formulation by conceptualizing the persistent scene state as dynamic ``fast weights'' \cite{schlag2021linear, schmidhuber1992learning} that are recursively updated during inference. However, while TTT3R employs an attention-based soft gating mechanism to modulate these updates, this heuristic approach lacks strict probabilistic bounds. Consequently, over long-horizon sequences, these unconstrained recursive updates inevitably suffer from progressive representational drift.

Mitigating such cumulative distortion requires a principled rethinking of state integration. On one hand, as evidenced by classical visual odometry \cite{jones2013accurate, williams2010combining} and recent video generation frameworks \cite{houanchor, liu2025anchorsync, zhang2025frame}, sequential drift can be explicitly bounded by anchoring recurrent updates to a reliable global reference (e.g., the initial frame), which prevents the latent state from deviating outside the valid geometric manifold. On the other hand, from a probabilistic standpoint, heuristic gating fundamentally deviates from the rigorous principles of Bayesian state estimation \cite{anderson2019chasing}. This uncertainty-aware paradigm has recently demonstrated profound success in suppressing drift across diverse vision tasks, including dynamic NeRFs \cite{zhan2024kfd}, streaming 3D Gaussians \cite{yan2025online}, and video diffusion \cite{yang2026confctrl}. By reformulating the update process within a Bayesian framework, the integration of novel information is no longer arbitrary; rather, it is strictly governed by an optimal balance between the prior covariance, which encodes the accumulated confidence of the historical state, and the measurement noise, which quantifies the intrinsic uncertainty of incoming observations.

\begin{figure}[t]
	\centering
	\includegraphics[width=\textwidth]{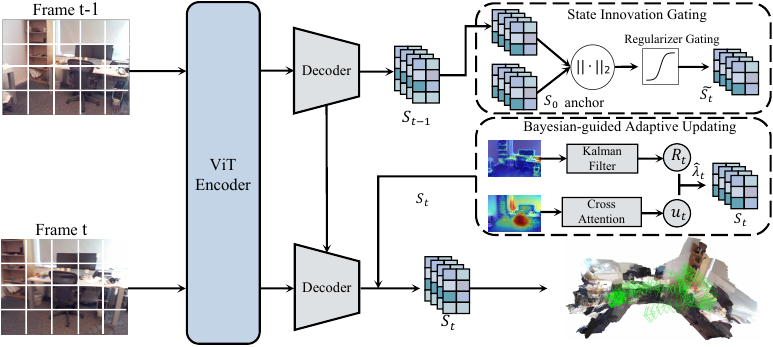}

	\caption{Overview of the BAT3R framework consisting of two modules State Innovation Gating (SIG) and Bayesian-guided Adaptive State Updating (BASU). SIG prevents representational drift by anchoring updates to the initial geometric state, while BASU optimizes observation fusion through a recursive Kalman filtering framework. By balancing observation reliability with historical confidence, the framework achieves robust, uncertainty-aware 3D scene reconstruction.}
	\label{main}
\end{figure}

While reformulating the recurrent update as Bayesian inference offers a principled solution to sequential drift, calibrating the requisite noise models conventionally demands domain-specific training. However, we observe that spatial and temporal uncertainty signals are already inherently encoded within the pretrained transformer. Specifically, cross-attention alignment implicitly captures spatial visibility, while the prediction head's confidence maps provide per-pixel proxies for observation reliability. Building upon this insight, we propose BAT3R, a training-free Bayesian adaptation framework for uncertainty-aware state updating in streaming 3D reconstruction, which comprises two synergistic modules (see Fig. \ref{main}). First, State Innovation Gating (SIG) acts as a distributional regularizer that explicitly anchors the recurrent state evolution to its geometric initialization, bounding spatial divergence by suppressing updates that fall outside the reliable latent manifold. Second, Bayesian-guided Adaptive State Updating (BASU) addresses optimal observation fusion by reformulating state integration as recursive Bayesian inference. By jointly modeling temporal observation reliability through information-theoretic gain and spatial relevance via cross-attention aggregation, BASU achieves an asymmetric spatio-temporal fusion, protecting well-consolidated memory from redundant updates while preserving full plasticity for novel scene content. Extensive experiments across seven standard benchmarks demonstrate that BAT3R substantially improves sequence length generalization. BAT3R reduces camera pose ATE by up to \textbf{67.8\%} and \textbf{32.3\%} relative to CUT3R and TTT3R, respectively, while maintaining the exact similar computational cost as the baselines.

\section{Related Works}
\subsection{Streaming Online Reconstruction Foundation Models.}
To improve scalability, streaming methods progressively register local geometries onto a canonical 3D space, mitigating computational overhead from redundant multi-view observations. StreamVGGT \cite{zhuo2025streaming} introduces a causal transformer for streaming 4D spatio-temporal reconstruction, employing temporal causal attention and KV-caching \cite{liu2024deepseek} as implicit memory for incremental historical context integration. However, it inherits the scaling bottlenecks of VGGT \cite{wang2025vggt}, incurring excessive resource overhead as frame count increases. STream3R \cite{yushi2026stream3r} reformulates 3D perception as a decoder-only task, leveraging causal attention for scalable reconstruction in both static and dynamic environments. Spann3R \cite{wang20253d} extends the DUSt3R paradigm with an external spatial memory to track 3D context, enabling direct regression of pointmaps into a unified global coordinate system. Point3R \cite{wu2025point3r} uses hierarchical 3D position embeddings and an explicit pointer-based memory to consolidate streaming visual information into a persistent scene representation. Recurrent architectures \cite{cabon2025must3r, chen2025long3r, wang20253d}, exemplified by CUT3R, maintain a persistent latent memory to reconcile historical context with new observations recursively. While they stabilize inference memory usage, the inherent forgetting problem \cite{chen2024stuffed, jelassi2024repeat} in recurrent units compromises long-term consistency, causing significant performance degradation as input views accumulate. More relevant models are provided in \ref{subsec:offline_foundation}.

\subsection{Test-Time Training for 3D Reconstruction.}
The $O(N^2)$ complexity of standard Transformers \cite{vaswani2017attention} has motivated the development of linear-complexity architectures. Modern RNNs and Linear Transformers \cite{katharopoulos2020transformers}, such as Mamba \cite{gu2024mamba, dao2024transformers} and DeltaNet \cite{schlag2021linear}, maintain a finite-sized hidden state as an efficient alternative. To counter information decay, these models incorporate advanced gating mechanisms, including input-dependent selection \cite{gu2024mamba} and delta-rule updates \cite{yang2024gated}, which regulate information flow across extended sequences. Paradigms such as Titans \cite{behrouz2024titans} and TTT \cite{sun2024learning, zhang2025test} reframe state updates as test-time optimization, enabling hidden states to function as fast weights \cite{sun2024learning, wang2025test} for rapid online adaptation \cite{liu2024longhorn}. Recent works have ported the TTT paradigm to 3D reconstruction, achieving linear-time inference by compressing historical context \cite{dherin2025learning} into fixed-capacity states. TTT3R \cite{chen2025ttt3r} reinterprets recurrent state update as an online learning problem, using test-time optimization to iteratively align historical geometric memory with incoming observations. ZipMap \cite{jin2026zipmap} employs a stateful architecture driven by test-time training to continuously compress dense pointmap priors, enabling scalable multi-view association across extended video streams. To mitigate catastrophic forgetting in recurrent frameworks, subsequent research has introduced training-free adaptive mechanisms. TTSA3R \cite{zheng2026ttsa3r} dynamically modulates update intensity based on temporal scene evolution, while MeMix \cite{dongmemix} partitions the recurrent state into independent patches to prevent transient noise from corrupting stable structural components. In contrast, BAT3R replaces heuristic updates with a principled Bayesian filter that mathematically models the confidence of spatial observations, achieving robust probabilistic state updates.

\section{Method}
\label{sec:method}

\subsection{Preliminaries}
TTT3R~\cite{chen2025ttt3r} builds on CUT3R~\cite{wang2025continuous} and treats streaming 3D reconstruction as test-time training: the recurrent state $\mathbf{S}_{t-1}\in\mathbb{R}^{N\times D}$ acts as fast weights~\cite{sun2024learning,schlag2021linear}, one vector per spatial token, updated at inference while the frozen encoder--decoder maps each frame. Each image $I_t$ and ray map $R_t$ are encoded into image tokens $\mathbf{X}_t\in\mathbb{R}^{M\times D}$ (with $M$ patches) and Plücker ray embeddings $P_t$, which are fed into the decoder together. A single recurrent decoder maps $(\mathbf{S}_{t-1},\mathbf{X}_t)$ through coupled streams: image-side tokens $\mathbf{Y}_t$ pass through a detokenizer to obtain the dense pointmap $\tilde{O}_t\in\mathbb{R}^{H\times W\times 3}$, while the state stream produces a novel state $\mathbf{S}_{t}$ aligned with the TTT interpretation of~\cite{chen2025ttt3r}. Abstractly, a test-time step subtracts a scaled pseudo-gradient from the previous fast weights:
\begin{equation}
	\mathbf{S}_{t} = \mathrm{Update}(\mathbf{S}_{t-1}, \mathbf{X}_t) \;=\; \mathbf{S}_{t-1} - \beta_t \, \nabla (\mathbf{S}_{t-1}, \mathbf{X}_t),
	\label{eq:ttt_general}
\end{equation}
where $\nabla (\mathbf{S}_{t-1}, \mathbf{X}_t)$ denotes the pseudo-gradient of the fast-weight update given the previous state and the current observation, and $\beta_t$ scales the descent step. TTT3R instantiates this term by cross-attention in which $\mathbf{S}_{t-1}$ supplies the queries and $\mathbf{X}_t$ supplies the keys and values:
\begin{equation}
	\nabla (\mathbf{S}_{t-1}, \mathbf{X}_t) \;=\; \operatorname{softmax}\!\left(\mathbf{Q}_{\mathbf{S}_{t-1}}\mathbf{K}_{\mathbf{X}_t}^{\top}\right)\mathbf{V}_{\mathbf{X}_t},
	\label{eq:ttt_grad_ca}
\end{equation}
where $\mathbf{Q}_{\mathbf{S}_{t-1}}=\mathbf{S}_{t-1}\mathbf{W}_Q$, $\mathbf{K}_{\mathbf{X}_t}=\mathbf{X}_t\mathbf{W}_K$, and $\mathbf{V}_{\mathbf{X}_t}=\mathbf{X}_t\mathbf{W}_V$ for learnable $\mathbf{W}_Q$, $\mathbf{W}_K$, and $\mathbf{W}_V$. Instead of using a fixed scalar for $\beta_t$, TTT3R derives a per-token gate $\boldsymbol{\beta}_t$ via an element-wise sigmoid applied to the update signal, merging the candidate state with the previous state via a gated exponential moving average:
\begin{equation}
	\boldsymbol{\beta}_t = \sigma\!\bigl(\nabla (\mathbf{S}_{t-1}, \mathbf{X}_t)\bigr), \qquad
	\mathbf{S}_t = (\mathbf{1}-\boldsymbol{\beta}_t)\odot \mathbf{S}_{t-1}+\boldsymbol{\beta}_t\odot \mathbf{S}_{t}.
	\label{eq:ttt_ema}
\end{equation}

Despite the soft-gating mechanism, the empirical updates in Eq. \eqref{eq:ttt_ema} lack explicit probabilistic bounds, leading to two critical bottlenecks over long horizons. First, without a global anchor, the unconstrained recursive updates cause the fast weights to progressively drift from their geometric initialization $\mathbf{S}_0$. Second, determining the gate $\boldsymbol{\beta}_t$ solely via spatial alignment ignores the intrinsic observation noise and historical state certainty, frequently overwriting well-consolidated memory.

\subsection{State Innovation Gating}
\label{sec:sig}
Error accumulation, as discussed above, stems from the absence of a fixed global reference in recurrent streaming, leading to a progressive departure of fast weights from the $\mathbf{S}_0$ initialization. Consequently, the latent state migrates outside the reliable regime of the frozen decoder, causing performance degradation. Validated by prior successes in visual odometry \cite{jones2013accurate} and video modeling \cite{liu2025anchorsync}, we propose \emph{State Innovation Gating} (SIG) to constrain state evolution. By anchoring each token of the recurrent state to the first-frame state $\mathbf{S}_0$, SIG measures the innovation residual and adaptively merges the decoder candidate $\mathbf{S}_t'$ with $\mathbf{S}_{t-1}$ via a gated convex combination.

For each token $i$, we formalize the innovation residual relative to the anchor $\mathbf{S}_0$ as:
\begin{equation}
	\delta_t^{(i)} =
	\bigl\|\mathbf{S}_t^{\prime(i)} - \mathbf{S}_0^{(i)}\bigr\|_2.
	\label{eq:sig_innovation}
\end{equation}
To maintain a consistent optimization landscape, we normalize this residual by the mean $\ell_2$ norm of the previous state, $r_t$, thereby avoiding scale inflation induced by recursive drift:
\begin{equation}
	r_t = \operatorname{mean}_i\bigl(\|\mathbf{S}_{t-1}^{(i)}\|_2\bigr).
	\label{eq:sig_scale}
\end{equation}
The adaptation process is modulated by a shifted sigmoid gate $g_t^{(i)}$:
\begin{equation}
	g_t^{(i)} = \operatorname{clip}\Bigl(
	\sigma\Bigl(-\kappa \cdot \tfrac{\delta_t^{(i)}}{r_t} + \tau\Bigr),\eta, 1.0\Bigr).
	\label{eq:sig_gate}
\end{equation}
Specifically, large deviations drive the gate toward its lower bound $\eta$, favoring the historical state $\mathbf{S}_{t-1}$ to curb divergence, whereas small ratios permit the candidate $\mathbf{S}_t'$ to pass with minimal modification. The final state $\widetilde{\mathbf{S}}_t$ is computed as:
\begin{equation}
	\widetilde{\mathbf{S}}_t =
	\mathbf{S}_t'\odot g_t + \mathbf{S}_{t-1}\odot(\mathbf{1}-g_t).
	\label{eq:sig_output}
\end{equation}
In the formulation, tokens that have drifted significantly from $\mathbf{S}_0$ are pulled back toward the previous state $\mathbf{S}_{t-1}$, while those remaining close to their initialization are permitted to pass through with minimal modification. The hyperparameter $\eta$ ensures that no token is completely frozen, maintaining a baseline level of plasticity. SIG requires no learnable parameters beyond these heuristic constraints, and its output $\widetilde{\mathbf{S}}_t$ serves as the refined observation for the subsequent BASU stage. The formal stability analysis and the proof of contraction toward the geometric anchor are detailed in \ref{subsec:theory_sig}.

\subsection{Bayesian-guided Adaptive State Updating}
\label{sec:brsf}
Given the drift-regularized candidate $\widetilde{\mathbf{S}}_t$, determining the optimal update magnitude for each state token becomes the central challenge. The heuristic gate $\beta_t$ in TTT3R relies exclusively on attention alignment, overlooking crucial factors: it neither assesses the intrinsic reliability of the prediction nor accounts for the accumulated confidence of the existing state token, and fails to distinguish whether a given token is even visible in the current frame. We address these two concerns jointly.

We augment the deterministic state with a per-token uncertainty $\boldsymbol{\Sigma}_t\in\mathbb{R}^{N}$, establishing a Gaussian belief $q_t=\mathcal{N}(\boldsymbol{\mu}_t, \mathrm{diag}(\boldsymbol{\Sigma}_t))$ over the $N$ state tokens. Here $\mathbf{S}_0\in\mathbb{R}^{N\times D}$ denotes the scene state obtained by encoding the first frame through the image encoder and projecting into the state space, and $\boldsymbol{\Sigma}_0=\sigma_0^2\mathbf{1}_N$ with $\sigma_0^2=10^3$, reflecting complete prior ignorance. $\boldsymbol{\Sigma}_t$ functions as a historical ledger of observation quality, progressively decreasing as reliable frames are accumulated over time.

To execute the Bayesian update, we must quantify both the unreliability of the incoming observation and the dynamic instability of the current state. First, we leverage the prediction head's confidence map $C_t$ as a data-driven proxy for per-pixel depth reliability, which is inversely mapped to a per-token observation noise covariance:
\begin{equation}
	\mathbf{R}_t = (C_t^\gamma+\epsilon)^{-1},
	\label{eq:obs_noise}
\end{equation}
so that well-textured regions yield low $\mathbf{R}_t$ (high observation confidence) and occluded or textureless areas yield high $\mathbf{R}_t$ (low observation confidence). The exponent $\gamma$ softens the confidence-to-noise mapping, preventing overly aggressive down-weighting of moderately uncertain observations. An adaptive process noise proportional to the innovation magnitude encodes inter-frame motion:
\begin{equation}
	\mathbf{Q}_t = e^{\alpha_Q}\cdot
	\bigl(1+\alpha_1\cdot\mathrm{softplus}(
	\|\widetilde{\mathbf{S}}_t-\boldsymbol{\mu}_{t-1}\|)\bigr),\quad
	\boldsymbol{\Sigma}_{t|t-1} = \boldsymbol{\Sigma}_{t-1}+\mathbf{Q}_t,
	\label{eq:proc_noise}
\end{equation}
so rapid motion inflates $\boldsymbol{\Sigma}_{t|t-1}$ and loosens the filter, while static scenes keep it tight. From $\boldsymbol{\Sigma}_{t|t-1}$ and $\mathbf{R}_t$, we compute the Kalman gain as the optimal fusion weight:
\begin{equation}
	\mathbf{K}_t = \omega\cdot
	\frac{\boldsymbol{\Sigma}_{t|t-1}}{\boldsymbol{\Sigma}_{t|t-1}+\mathbf{R}_t}
	+(1-\omega)\cdot
	\frac{\sqrt{\boldsymbol{\Sigma}_{t|t-1}}}{\sqrt{\boldsymbol{\Sigma}_{t|t-1}}
		+\sqrt{\mathbf{R}_t}},
	\label{eq:kalman_gain}
\end{equation}
where $\omega\in[0,1]$ is a weight that interpolates between the standard ratio form term and a square-root term. The standard ratio naturally follows from Gaussian Bayesian inference, while the square-root variant offers improved numerical stability when $\boldsymbol{\Sigma}_{t|t-1}$ and $\mathbf{R}_t$ differ by orders of magnitude. $\mathbf{K}_t$ drives aggressive updates ($\mathbf{K}_t\to 1$) only when the state is highly uncertain and the observation is reliable; otherwise it conservatively preserves the existing memory. The posterior updates:
\begin{align}
	\boldsymbol{\mu}_{t|t} &=
	\boldsymbol{\mu}_{t-1}+\mathbf{K}_t\cdot
	(\widetilde{\mathbf{S}}_t-\boldsymbol{\mu}_{t-1}),
	\label{eq:mu_update}\\
	\boldsymbol{\Sigma}_{t|t} &=
	(\mathbf{I}-\mathbf{K}_t)\boldsymbol{\Sigma}_{t|t-1},\quad
	\boldsymbol{\Sigma}_{t|t}\geq\rho\,\boldsymbol{\Sigma}_{t|t-1}.
	\label{eq:sigma_update}
\end{align}
Each reliable integration reduces $\boldsymbol{\Sigma}_{t|t}$ toward its floor, making well-observed tokens progressively more resistant to future noisy updates. We provide theoretical justifications for the uncertainty boundedness and the optimality of this Bayesian fusion process in \ref{subsec:theory_manifold} and \ref{subsec:theory_bayesian}, respectively. We quantify the information gain of an update by computing the symmetric Kullback-Leibler (KL) divergence between the posterior and the prior:
\begin{equation}
	\mathrm{IG}_t =
	\tfrac{1}{2}\Bigl[
	\tfrac{\boldsymbol{\Sigma}_{t|t-1}}{\boldsymbol{\Sigma}_{t|t}}+
	\tfrac{\boldsymbol{\Sigma}_{t|t}}{\boldsymbol{\Sigma}_{t|t-1}}-2\Bigr]
	+\tfrac{(\boldsymbol{\mu}_{t|t}-\boldsymbol{\mu}_{t-1})^2}{2}
	\bigl(\boldsymbol{\Sigma}_{t|t-1}^{-1}+\boldsymbol{\Sigma}_{t|t}^{-1}\bigr).
	\label{eq:ig}
\end{equation}
The posterior statistics are squashed into a per-token reliability gate $\lambda_t\in(0,1)$ via a pre-calibrated sigmoid function. Consequently, updates that are anomalously large (likely outliers) or negligibly small (redundant observations) are severely penalized, whereas genuinely informative updates are substantially integrated. However, even high-fidelity observations are strictly bounded by the current camera frustum; applying the temporal Bayesian gain $\lambda_t$ uniformly across the entire state risks corrupting well-consolidated memory residing outside the active field of view. To circumvent this, we modulate $\lambda_t$ with a spatial relevance weight derived from the decoder's intrinsic cross-attention maps. Since the recurrent state assimilates novel geometry by querying incoming visual tokens, these attention affinities inherently capture spatial visibility. Formally, for a given state token $i$, we compute an aggregated visibility score $a_t^{(i)}$ by accumulating its cross-attention probabilities to all incoming image tokens across all $L$ layers and $H$ heads. The score is subsequently projected into a normalized spatial relevance weight $\mathbf{u}_t^{(i)}=\sigma(a_t^{(i)})\in[0,1]$. Ultimately, the temporal reliability gate and the spatial relevance weight are synergized into a unified update mask:
\begin{equation}
	\hat{\lambda}_t = \lambda_t\odot(\alpha + (1-\alpha)\cdot\mathbf{u}_t),
	\label{eq:agsu_gate}
\end{equation}
where $\alpha \in [0,1]$ preventing complete token exclusion. The state is finally formulated as:
\begin{equation}
	\mathbf{S}_t = \hat{\lambda}_t\odot\boldsymbol{\mu}_{t|t}
	+(\mathbf{1}-\hat{\lambda}_t)\odot\mathbf{S}_{t-1}.
	\label{eq:agsu_write}
\end{equation}
Tokens with high temporal reliability and high spatial coverage receive the strongest Bayesian update, whereas tokens that are either unreliably observed or outside the current frustum are safely preserved from $\mathbf{S}_{t-1}$. To close the recursive loop, the state covariance is updated as $\boldsymbol{\Sigma}_t=\hat{\lambda}_t \boldsymbol{\Sigma}_{t|t}+(1-\hat{\lambda}_t)\boldsymbol{\Sigma}_{t|t-1}$ and carried forward.

\begin{figure}[t]
	\centering

	\begin{subfigure}[b]{0.48\linewidth}
		\centering
		\includegraphics[width=\textwidth]{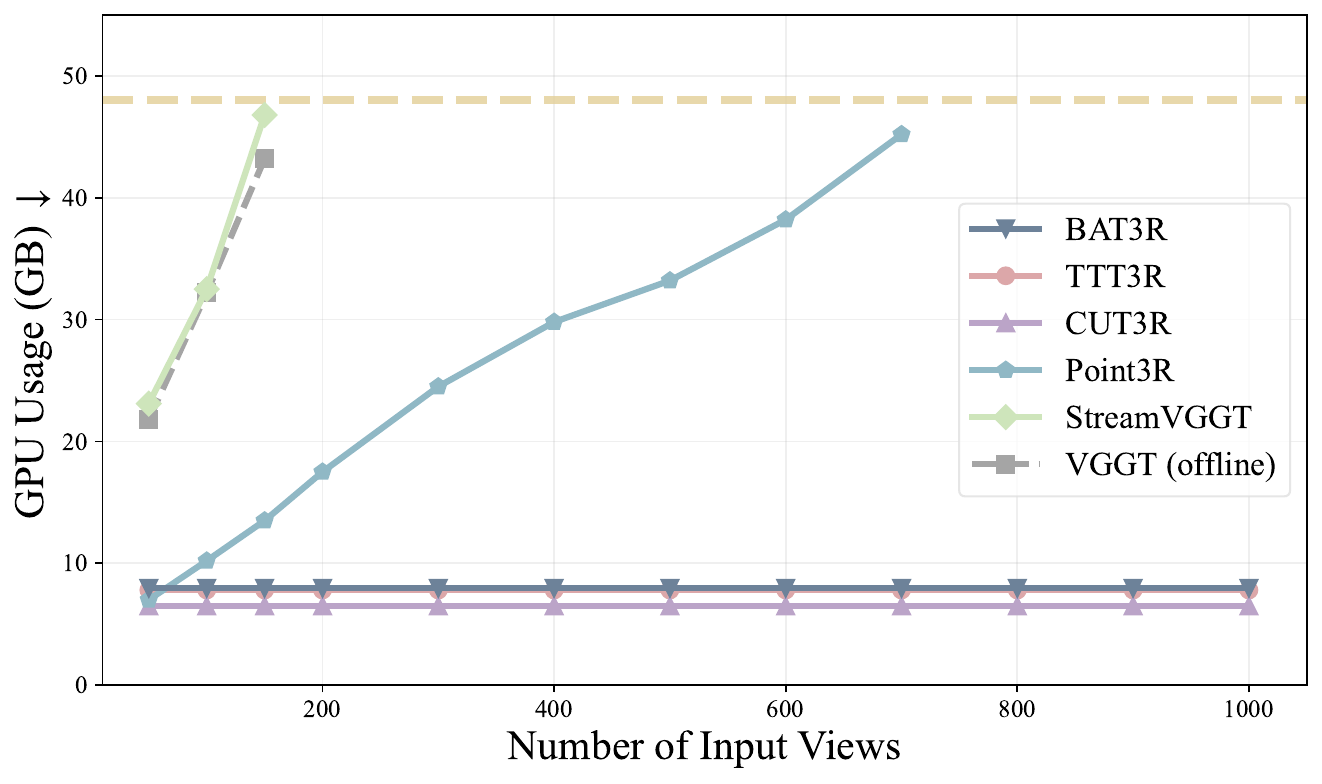}
		\caption{Comparison of GPU memory consumption (GB)}
		\label{fig:memory}
	\end{subfigure}
	\hfill

	\begin{subfigure}[b]{0.48\linewidth}
		\centering
		\includegraphics[width=\textwidth]{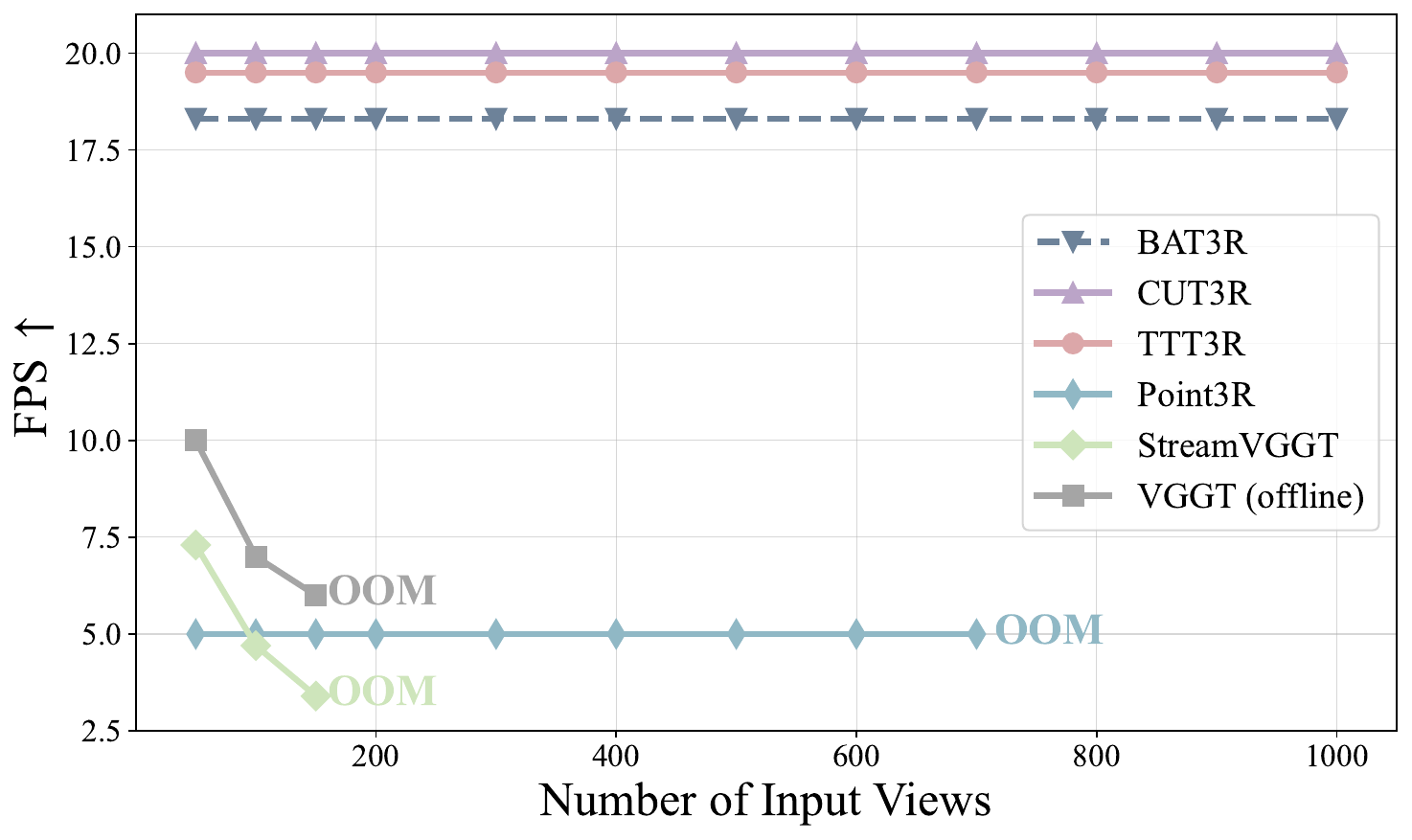}
		\caption{Comparison of processing speed (FPS)}
		\label{fig:fps}
	\end{subfigure}

	\caption{\textbf{Evaluation of computational efficiency on ScanNet.} (a) GPU memory and (b) inference speed across varying numbers of input views.}
	\label{fig:efficiency_scannet}
\end{figure}

\section{Experiments}
\subsection{Experimental Setup}
\textbf{Datasets.} We evaluate BAT3R on three tasks across Seven benchmarks.
For camera pose estimation, we evaluate BAT3R on TUM dynamics \cite{sturm2012benchmark}, ScanNet \cite{dai2017scannet}, and Sintel \cite{butler2012naturalistic} datasets. TUM dynamics contains indoor RGB-D sequences captured with a hand-held camera, including both static scenes and dynamic scenes with moving people. ScanNet contains large-scale indoor RGB-D sequences captured across diverse room types with significant viewpoint variation. Sintel contains long synthetic sequences rendered from a 3D animated film, featuring complex motions, dynamic objects, and challenging lighting conditions. For video depth estimation, we evaluate BAT3R on Bonn \cite{palazzolo2019refusion}, KITTI \cite{geiger2013vision}, and Sintel \cite{butler2012naturalistic} datasets. Bonn contains highly dynamic indoor RGB-D sequences, featuring complex human activities, fast-moving objects, and varying camera motions. KITTI contains large-scale outdoor driving sequences captured from a moving vehicle, including real-world street views with dynamic entities such as pedestrians and cars. For 3D reconstruction, we evaluate on 7-Scenes \cite{shotton2013scene} and NRGBD \cite{azinovic2022neural}, which contains indoor RGB-D sequences of seven distinct room categories captured with a Kinect sensor, commonly used to assess geometric reconstruction accuracy.

\textbf{Baselines}. We evaluate BAT3R against state-of-the-art online 3D reconstruction baselines, including the recurrent CUT3R \cite{wang2025continuous} and its long-sequence extensions, Point3R \cite{wu2025point3r} and StreamVGGT \cite{zhuo2025streaming}, which rely on explicit spatial memory and KV-caching mechanisms. Furthermore, we compare against recent training-free Test-Time Training (TTT) frameworks , such as TTT3R \cite{chen2025ttt3r} and TTSA3R \cite{zheng2026ttsa3r}, which utilize heuristic gating and spatio-temporal alignment for adaptive state updates.

\subsection{Camera Pose Estimation}
\begin{figure*}[htbp]
	\centering
	\includegraphics[width=\textwidth]{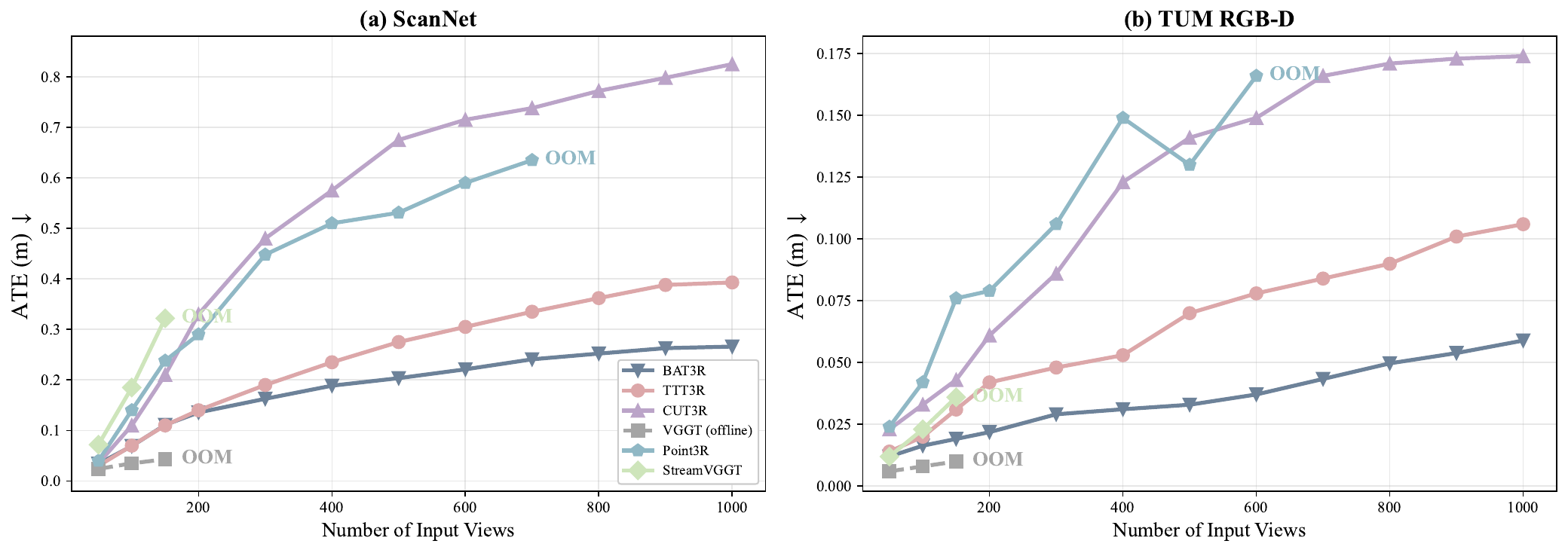}
	\caption{Comparison of Camera Pose Estimation. Results on ScanNet (left) and TUM-D (right). OOM denotes the method out-of-memory beyond this point.}
	\label{fig:pose_dataset}
\end{figure*}

Consistent with established evaluation protocols \cite{zhang2024monst3r}, we assess camera pose estimation on the TUM dynamics \cite{sturm2012benchmark} and ScanNet \cite{dai2017scannet} datasets. Performance is measured using the standard Absolute Translation Error (ATE) following Sim(3) trajectory alignment \cite{umeyama2002least} between the estimated and ground-truth trajectories. The long-sequence evaluation results are presented in Fig. \ref{fig:pose_dataset}. For a fair comparison, we adopt the experimental settings and baseline data established by TTT3R \cite{chen2025ttt3r}. Empirical evaluations demonstrate that BAT3R consistently bounds representational drift, sustaining high-accuracy pose estimation across widely varying sequence lengths ranging from 50 to 1000 frames. When scaling to 1000-frame sequences, BAT3R reduces ATE by 67.8\% on ScanNet and 66.1\% on TUM dynamics compared to the CUT3R baseline. Even when compared to the competitive TTT3R, BAT3R achieves a further error reduction of 32.3\% and 44.4\% on ScanNet and TUM, respectively. Besides, we also conduct comprehensive evaluations of GPU memory consumption and inference speed on the ScanNet dataset in Fig. \ref{fig:efficiency_scannet}. As is shown in Fig. \ref{fig:memory}, \ref{fig:fps},  BAT3R achieves competitive GPU memory and inference speed compared with previous methods. Additional quantitative comparisons against offline methods on short sequences and qualitative visualizations of the estimated camera trajectories can be found in \ref{subsec:pose_eval} and \ref{subsec:depth_eval}.

\subsection{Video Depth Estimation}
\begin{figure*}[htbp]
	\centering
	\includegraphics[width=\textwidth]{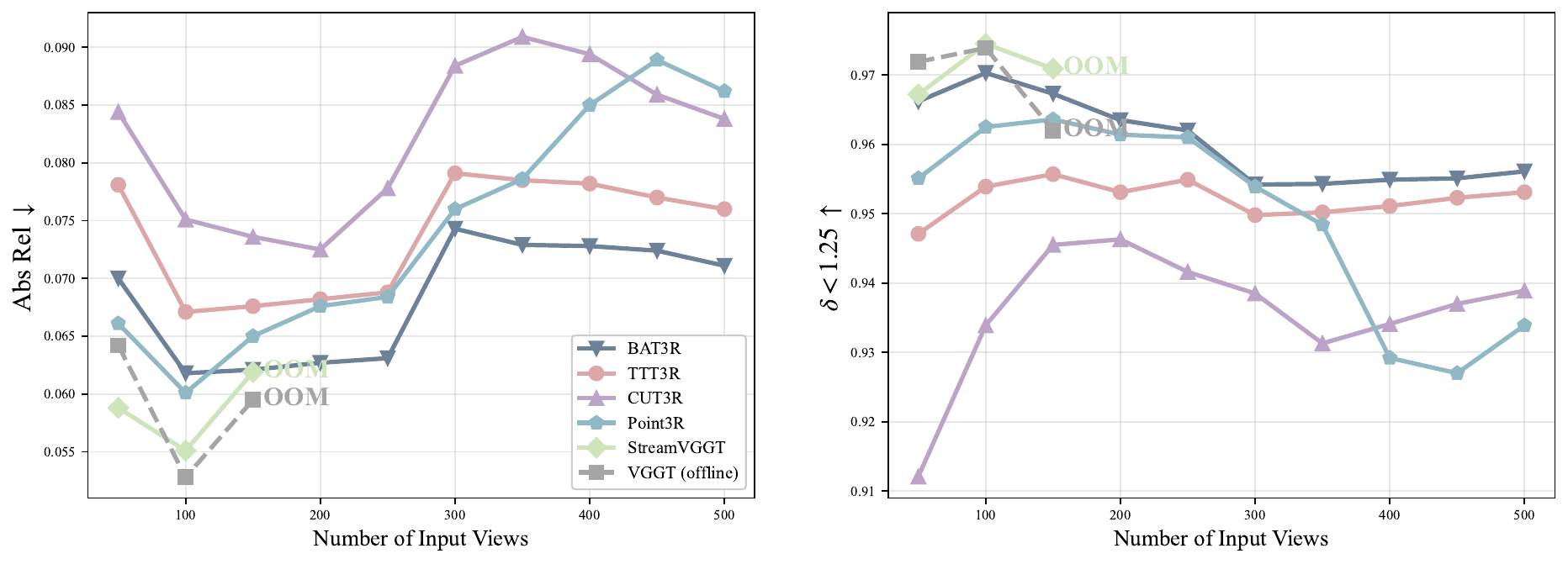}
	\caption{Scale-invariant relative depth evaluation on Bonn dataset.}
	\label{fig:bonn_dataset}
\end{figure*}

Following standard evaluation protocols \cite{zhang2024monst3r}, we assess video depth estimation on the KITTI \cite{geiger2013vision} and Bonn \cite{palazzolo2019refusion} datasets, which encompass a variety of indoor and outdoor driving environments. We adopt Absolute Relative Error (Abs Rel) and $\delta < 1.25$ as our primary performance metrics. Following the evaluation protocol established in TTT3R \cite{chen2025ttt3r}, relative depth accuracy and temporal consistency are gauged by aligning predictions to the ground truth using a per-sequence scale, while metric-scale performance is assessed directly. As illustrated in the long-sequence evaluations in Fig. \ref{fig:bonn_dataset} and Fig. \ref{fig:kitti_dataset}, BAT3R method consistently outperforms existing recurrent baselines. On the Bonn dataset, BAT3R achieves a substantial improvement, reducing the Abs Rel by approximately 15.5\% compared to the CUT3R baseline at a sequence length of 500 frames. Similarly, on the KITTI dataset, BAT3R effectively bounds progressive state drift, yielding a 24.5\% and 13.6\% reduction in Abs Rel relative to the CUT3R and TTT3R baselines at a sequence length of 500 frames. Furthermore, as the sequence length increases, BAT3R mitigates the temporal degradation in the $\delta < 1.25$ metric, sustaining a more stable and consistently higher $\delta < 1.25$ across all sequence lengths.

\begin{figure*}[htbp]
	\centering
	\includegraphics[width=\textwidth]{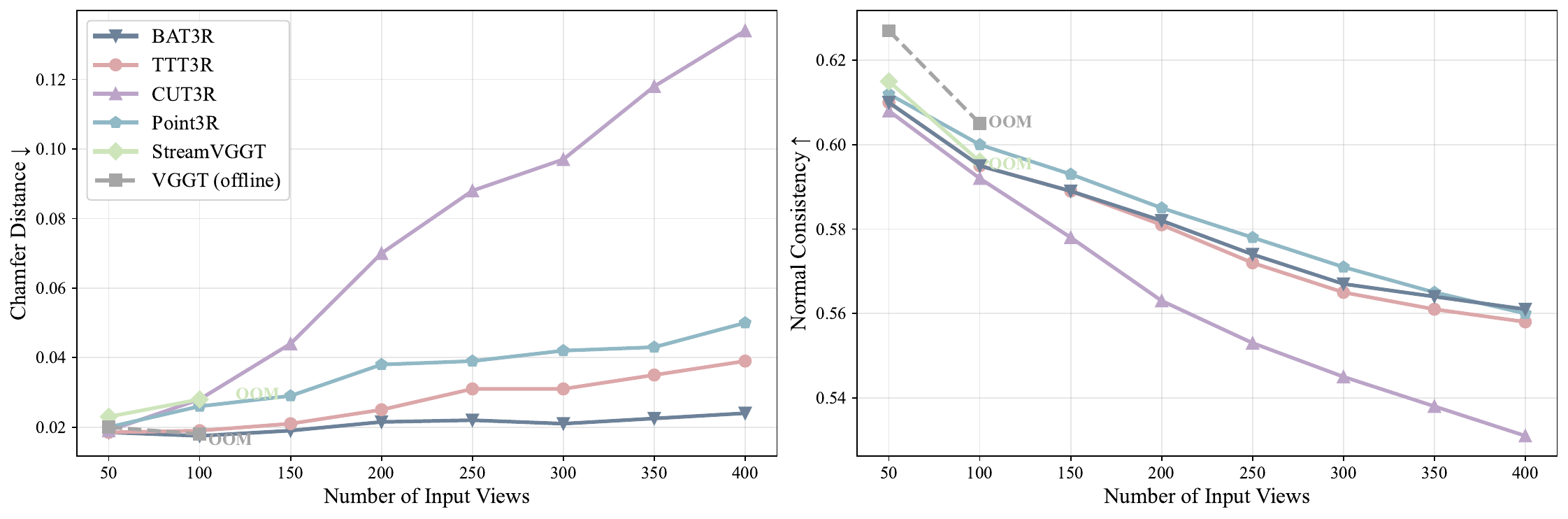}
	\caption{Comparison of 3D Reconstruction on 7-scene.}
	\label{fig:7scenes_dataset}
\end{figure*}

\subsection{3D Reconstruction}
\begin{figure*}[t]
	\centering

	\begin{tabular}{@{}c@{\hspace{2pt}}ccc@{}}

		\begin{sideways}\quad \small \textbf{BAT3R}\end{sideways} &
		\includegraphics[width=0.3\textwidth, height=2.5cm, keepaspectratio]{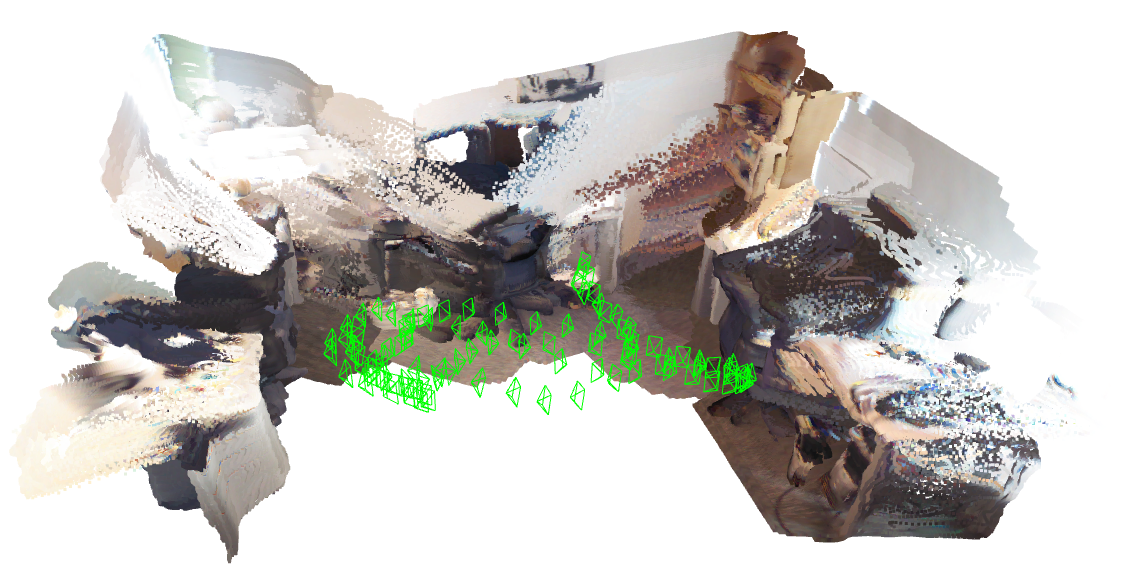} &
		\includegraphics[width=0.3\textwidth, height=2.5cm, keepaspectratio]{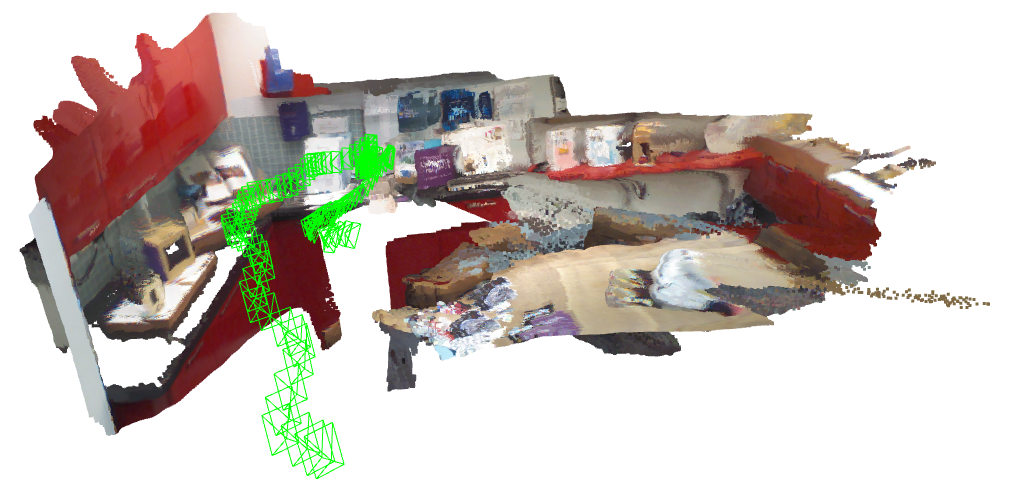} &
		\includegraphics[width=0.3\textwidth, height=2.5cm, keepaspectratio]{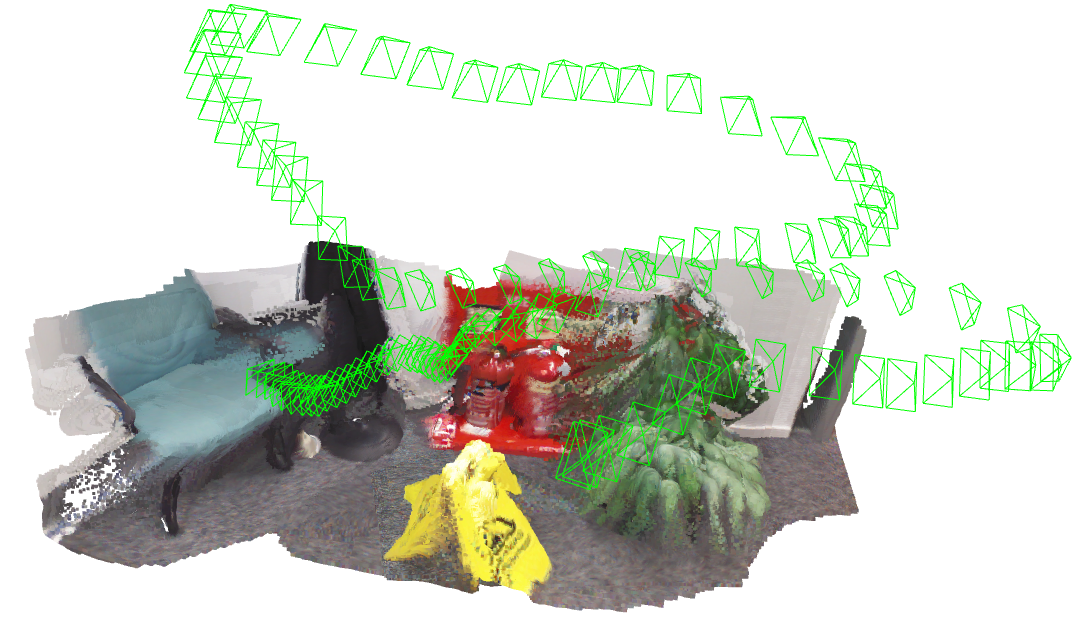} \\

		\begin{sideways}\quad \small \textbf{TTT3R}\end{sideways} &
		\includegraphics[width=0.3\textwidth, height=2.5cm, keepaspectratio]{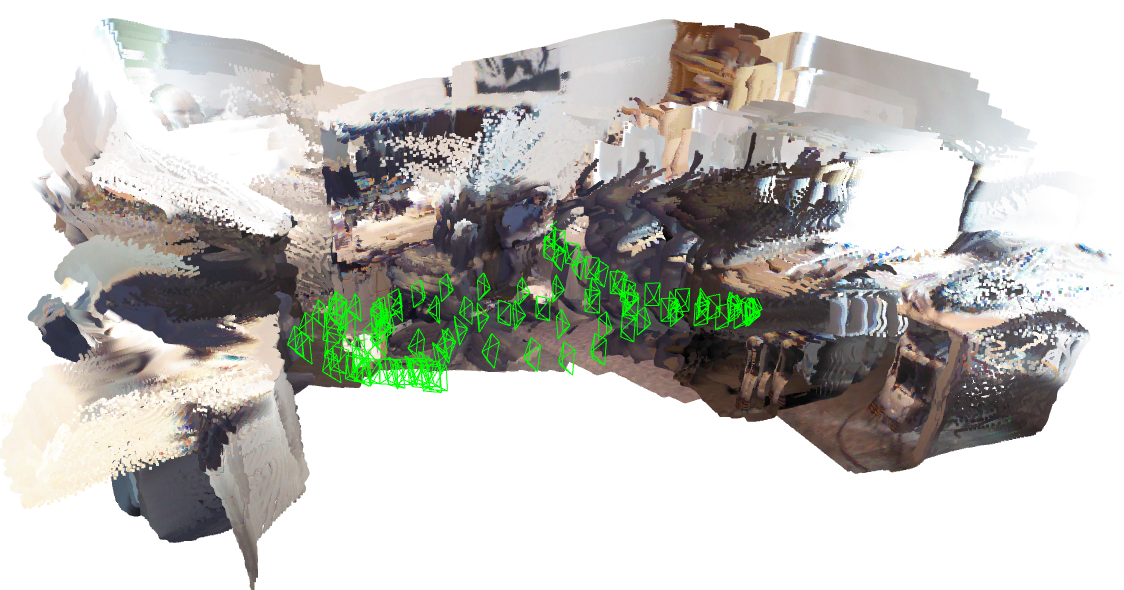} &
		\includegraphics[width=0.3\textwidth, height=2.5cm, keepaspectratio]{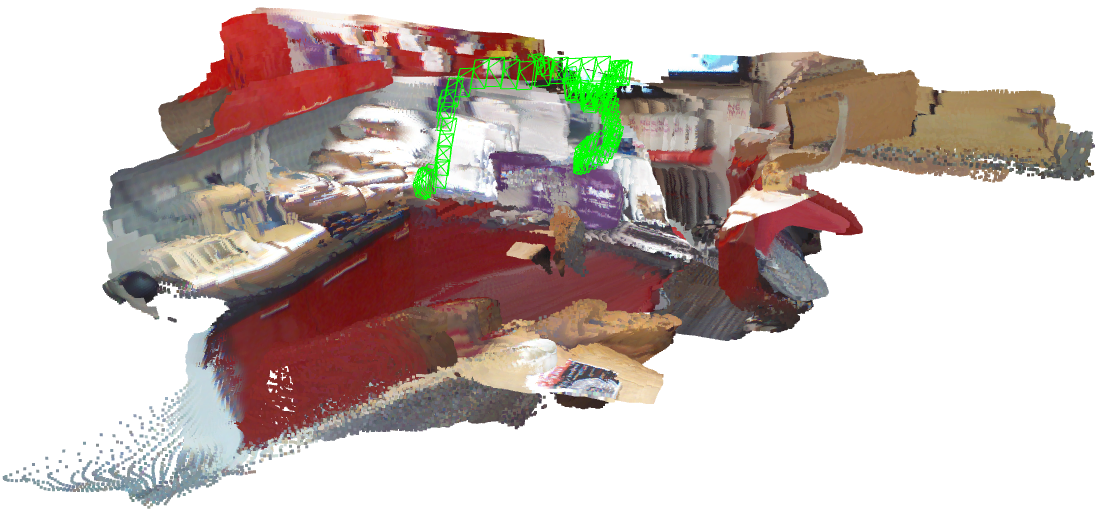} &
		\includegraphics[width=0.3\textwidth, height=2.5cm, keepaspectratio]{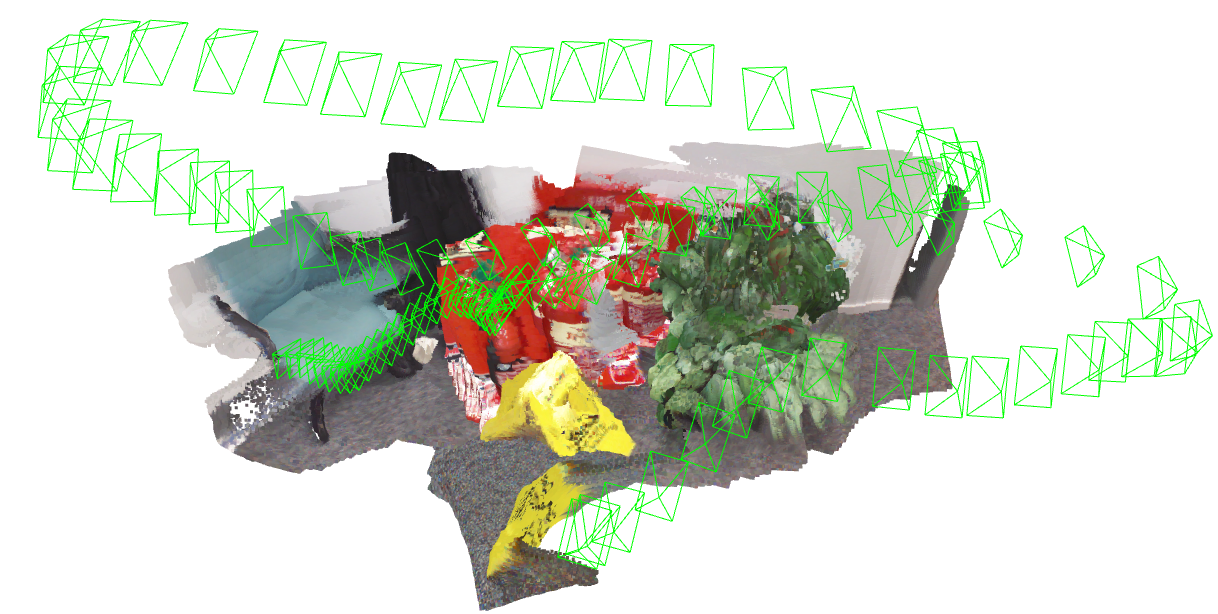} \\

		\begin{sideways}\quad \small \textbf{CUT3R}\end{sideways} &
		\includegraphics[width=0.3\textwidth, height=2.5cm, keepaspectratio]{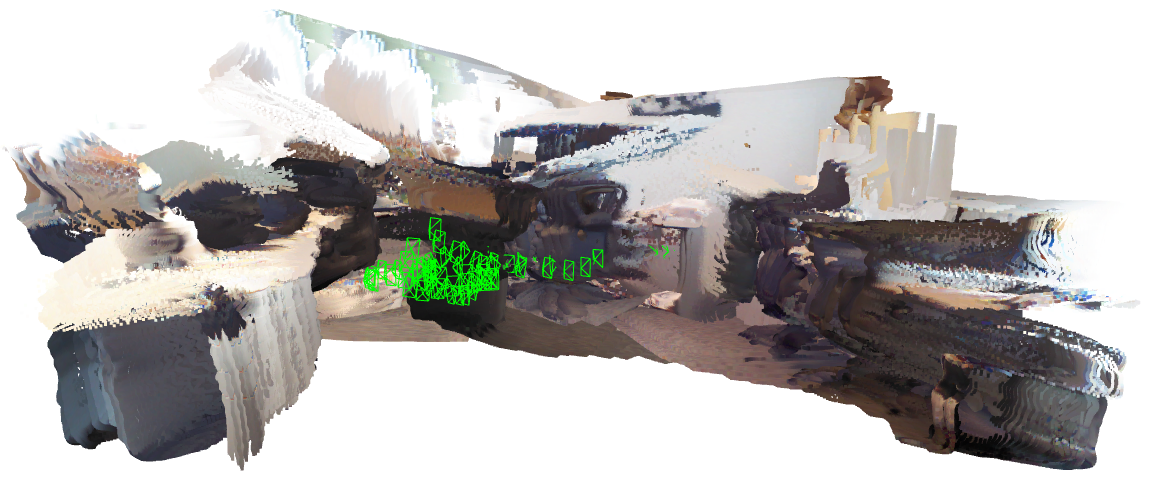} &
		\includegraphics[width=0.3\textwidth, height=2.5cm, keepaspectratio]{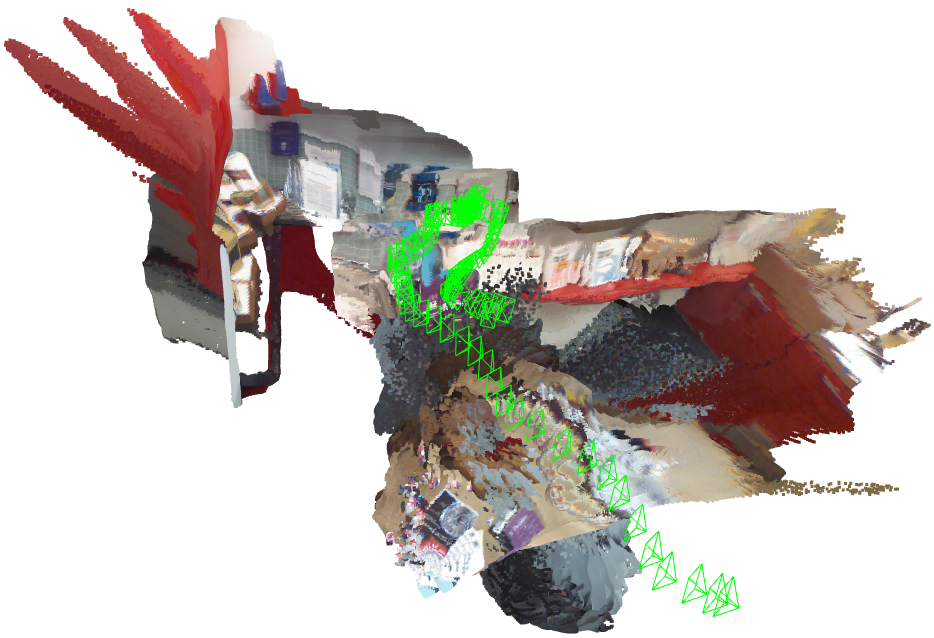} &
		\includegraphics[width=0.3\textwidth, height=2.5cm, keepaspectratio]{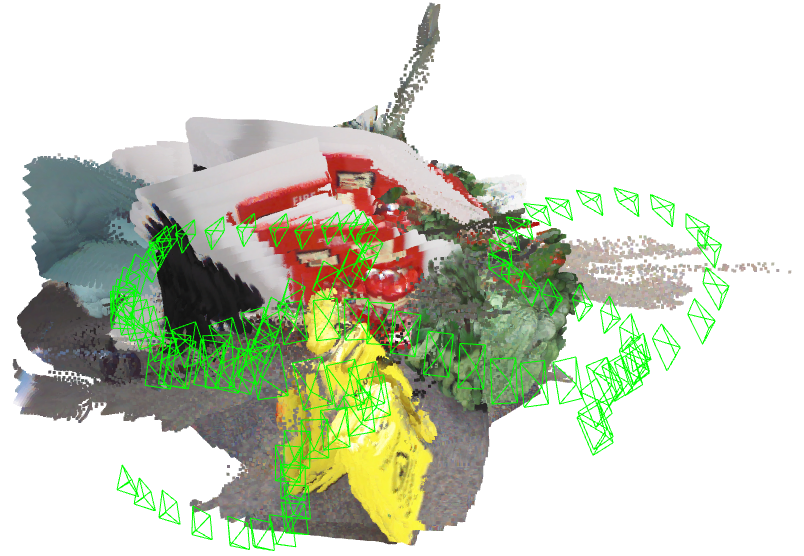} \\
	\end{tabular}

	\caption{\textbf{Qualitative comparison of 3D reconstruction.} We compare BAT3R with TTT3R and CUT3R across three challenging sequences in the  7-Scene dataset. }
	\label{fig:qualitative_comparison}
\end{figure*}

Following established protocols \cite{wang20253d, wang2025continuous, chen2025feat2gs}, we evaluate 3D reconstruction on the 7-Scenes \cite{shotton2013scene} dataset. Geometric accuracy is quantified through Chamfer Distance which denotes the average of accuracy and completeness, and Normal Consistency (NC). In contrast to the sparse sampling, we conduct evaluations on extended image sequences to rigorously assess the models' capacity for long-term spatial memorization. As illustrated in the long-sequence evaluations in Fig. \ref{fig:7scenes_dataset}, our BAT3R method consistently outperforms existing recurrent baselines. On the 7-Scenes benchmark, BAT3R achieves a substantial improvement, reducing the Chamfer Distance (from 0.134 to 0.024) by approximately 82.1\% compared to the CUT3R baseline at a sequence length of 400 frames. Furthermore, when compared to the highly competitive TTT3R, BAT3R achieves a further reduction of approximately 38.5\% (from 0.039 to 0.024) while maintaining a slight but consistent lead in Normal Consistency (e.g., 0.561 vs. 0.558). We provide qualitative comparisons of these reconstruction results in Fig. \ref{fig:qualitative_comparison}. Furthermore, the detailed quantitative results on the 7-Scenes and NRGBD benchmarks across various sequence lengths are provided in \ref{subsec:recon_eval}, demonstrating that BAT3R consistently yields substantial error reductions compared to SOTA baselines (TTSA3R, TTT3R, and CUT3R).

\begin{table*}[htbp]
	\centering
	\caption{Comprehensive Ablation Study on Camera Pose and Video Depth Estimation. We evaluate the impact of different components across four standard benchmarks.}
	\label{tab:ablation_overall}
	\resizebox{\textwidth}{!}{
		\begin{tabular}{l ccc ccc cccc}
			\toprule
			\multirow{3}{*}{\textbf{Setting}} & \multicolumn{6}{c}{\textbf{Camera Pose Estimation}} & \multicolumn{4}{c}{\textbf{Video Depth Estimation}} \\
			\cmidrule(lr){2-7} \cmidrule(lr){8-11}
			& \multicolumn{3}{c}{\textbf{TUM-dynamics (90 frames)}} & \multicolumn{3}{c}{\textbf{ScanNet (90 frames)}} & \multicolumn{2}{c}{\textbf{Bonn (100 frames)}} & \multicolumn{2}{c}{\textbf{KITTI (100 frames)}} \\
			\cmidrule(lr){2-4} \cmidrule(lr){5-7} \cmidrule(lr){8-9} \cmidrule(lr){10-11}
			& ATE $\downarrow$ & RPE trans $\downarrow$ & RPE rot $\downarrow$ & ATE $\downarrow$ & RPE trans $\downarrow$ & RPE rot $\downarrow$ & Abs Rel $\downarrow$ & $\delta < 1.25 \uparrow$ & Abs Rel $\downarrow$ & $\delta < 1.25 \uparrow$ \\
			\midrule
			BAT3R (Full)   & \textbf{0.016} & \textbf{0.008} & \textbf{0.287} & \textbf{0.058} & \textbf{0.022} & \textbf{0.852} & \textbf{0.061} & \textbf{97.0} & \textbf{0.100} & \textbf{89.2} \\
			\midrule
			w/o BRU        & 0.024 & 0.008 & 0.288 & 0.099 & 0.021 & 0.562 & 0.062 & 96.5 & 0.106 & 87.9 \\
			w/o SIG        & 0.016 & 0.008 & 0.289 & 0.061 & 0.025 & 1.163 & 0.065 & 95.6 & 0.105 & 88.8 \\
			w/o Attention  & 0.017 & 0.008 & 0.260 & 0.060 & 0.024 & 0.988 & 0.063 & 96.8 & 0.105 & 88.1 \\
			w/o Q          & 0.017 & 0.009 & 0.300 & 0.077 & 0.028 & 1.206 & 0.061 & 96.7 & 0.104 & 88.5 \\
			\bottomrule
		\end{tabular}
	}
\end{table*}

\subsection{Ablation Studies}
\textbf{Effect of Each Component.}
To evaluate the individual components of BAT3R, we conduct comprehensive ablation studies on camera pose and depth estimation (Table \ref{tab:ablation_overall}). Removing the Bayesian Recursive Update (BRU) causes the most severe degradation (ScanNet ATE surging from 0.058 to 0.099), validating its necessity in optimally balancing historical confidence against observation noise. Omitting the State Innovation Gating (SIG) increases Abs Rel from 0.061 to 0.065 on Bonn, confirming its critical role as a distributional regularizer to bound representational drift. Furthermore, excluding cross-attention compromises spatial awareness, dropping the KITTI $\delta < 1.25$ metric from 89.2 to 88.1. Finally, the absence of adaptive process noise ($\mathbf{Q}_t$) degrades performance (ATE increasing to 0.077), verifying that motion-aware uncertainty dynamically prevents the Kalman filter from becoming overly restrictive.

\textbf{Parameter Analysis.} Furthermore, we conduct hyperparameter sensitivity experiments on the TUM-dynamics dataset to assess the impact of the update floor $\eta$, scaling factor $\kappa$, and threshold $\tau$. The detailed experimental results and analysis are provided in Appendix. As illustrated in Fig. \ref{fig:overall_ablation}, the Average ATE is least sensitive to the floor $\eta$, whereas $\kappa$ and $\tau$ both exhibit a sharp initial performance gain that plateaus once they exceed $1.0$ and $5.0$, respectively. Furthermore, additional exhaustive analysis of both SIG and BASU hyperparameters, including the scaling factors for process noise and interpolation weights, is detailed in \ref{subsec:param_analysis}.

\section{Conclusion}
In this work, we present BAT3R, a training-free Bayesian adaptation framework that mitigates catastrophic forgetting and representational drift in streaming 3D reconstruction. BAT3R comprises two synergistic modules: State Innovation Gating (SIG), which acts as a distributional regularizer to anchor state evolution to its geometric initialization, and Bayesian-guided Adaptive State Updating (BASU), which reformulates state integration as a recursive Bayesian inference process within a Kalman filtering framework. By dynamically balancing the reliability of incoming observations against the accumulated confidence of historical memory, BAT3R achieves an asymmetric spatio-temporal state fusion that protects well-consolidated regions from redundant updates while preserving plasticity for genuinely novel content. Extensive experiments across seven standard benchmarks demonstrate consistent improvements over state-of-the-art online methods, with particularly strong gains on long sequences. BAT3R achieves these results without any training or fine-tuning, making it immediately applicable to any pretrained streaming reconstruction model.

\section*{Acknowledgments}
This work was supported in part by the Science and Technology Development Fund, Macau SAR, under Grant 0193/2023/RIA3, 0079/2025/AFJ and 0052/2026/RIA1.

\bibliographystyle{plainnat}
\bibliography{references}

\clearpage

\appendix

\clearpage
\appendix

\phantomsection
\label{sec:appendix}

\noindent{\Huge \textbf{Appendix}}
\vspace{1cm}

\vspace{0.2cm}

\startcontents[appendix]
\printcontents[appendix]{l}{1}{\setcounter{tocdepth}{2}}
\clearpage

\section{Related Works}
\label{sec:supp_related_works}
\subsection{Offline Reconstruction Foundation Models.}
\label{subsec:offline_foundation}
Recent advances in 3D vision have been driven by feed-forward foundation models that streamline the traditional reconstruction pipeline. DUSt3R \cite{wang2024dust3r} pioneers this shift with an end-to-end framework that directly regresses pixel-aligned pointmaps \cite{brachmann2017dsac, brachmann2024scene, shotton2013scene} from image pairs, bypassing explicit feature matching \cite{an2025cross, chen2025easi3r} and camera calibration. MASt3R \cite{leroy2024grounding} extends this with cross-view correspondences for 3D matching, though it remains limited to pairwise inputs and requires heavy global optimization for multi-view integration. Fast3R \cite{yang2025fast3r} leverages global self-attention to jointly reason across all input frames in a single forward pass, achieving strong accuracy in camera pose and 3D reconstruction without iterative alignment. VGGT \cite{wang2025vggt} introduces a feed-forward framework that reconstructs 3D attributes from an arbitrary number of views in one pass. $\pi^3$ \cite{wang2025pi} employs a permutation-equivariant architecture to predict camera poses and point maps in a reference-free manner, ensuring robustness to input ordering. Despite inference acceleration through token merging or sparse attention \cite{shen2025fastvggt}, the $O(N^2)$ complexity of full attention remains a fundamental bottleneck: it routinely causes out-of-memory errors on consumer GPUs, limiting the maximum sequence length that can be processed.

\section{Additional Results}
\label{sec:additional_results}
\subsection{Experimental Settings}
\label{subsec:exp_settings}
\textbf{Evaluation Settings.} To thoroughly evaluate our model, we establish two primary evaluation setups: (1) \textit{Long-sequence evaluation} (results in the main paper), which challenges the recurrent state's ability to maintain long-term memory and geometric consistency over hundreds of frames against other online methods. (2) \textit{Short-sequence evaluation} (detailed in this supplement), which allows for benchmarking against offline or memory-intensive baselines that would otherwise fail due to out-of-memory (OOM) errors on longer videos.

\textbf{Implementation Details.} In all experiments, inference is conducted on a single NVIDIA RTX 4090 GPU (24 GB). BAT3R is built upon the pretrained CUT3R checkpoint and operates strictly without any fine-tuning or additional training. The underlying architecture consists of a ViT-Large encoder (24 layers, embedding dimension of 1024, 16 attention heads) coupled with a 12-layer recurrent decoder (embedding dimension of 768, 12 attention heads). At the beginning of each sequence, the Bayesian state variance is initialized to $1e^3$ per patch, reflecting a state of high initial uncertainty. Consistent with prior training-free adaptation frameworks \cite{chen2025ttt3r, zheng2026ttsa3r}, we do not perform any test-time gradient updates; instead, continuous state adaptation is achieved entirely through our proposed Bayesian state update rules.
\subsection{Baselines}
\label{subsec:baselines}
We first compare BAT3R against pairwise 3D reconstruction foundation models (e.g., DUSt3R \cite{wang2024dust3r} , MASt3R \cite{leroy2024grounding}, MonST3R \cite{zhang2024monst3r}, and Easi3R \cite{chen2025easi3r}), which take view pairs as input and require an extra global alignment stage to integrate their predictions. We also evaluate against AETHER \cite{zhu2025aether}  and VGGT \cite{wang2025vggt}, which simultaneously predict all pointmaps to achieve state-of-the-art 3D point and camera pose reconstruction. However, constrained by their reliance on global alignment or full-attention mechanisms, these methods are restricted to offline processing of short sequences, necessitating the re-inference of the entire sequence whenever a new frame arrives.

For online methods, we compare BAT3R with Spann3R \cite{wang20253d} and CUT3R \cite{wang2025continuous}, which leverage RNN-based architectures to process streaming images on the fly. Additionally, we compare against the most relevant concurrent works, Point3R \cite{wu2025point3r} and StreamVGGT \cite{zhuo2025streaming}. To handle long sequences, these approaches fine-tune CUT3R and VGGT by introducing explicit pointmap memory and KV caching as their respective state representations. In contrast to these methods, TTT3R reformulate CUT3R from a Test-Time Training (TTT) perspective. Build on TTT3R, BAT3R achieves robust online perception by integrating State Innovation Gating (SIG) to constrain representational drift and Bayesian-guided Adaptive State Updating (BASU) for optimal, uncertainty-aware fusion , entirely eliminating the need to fine-tune CUT3R or train additional parameterized components.

\begin{table*}[hbp]
	\centering
	\caption{\textbf{Evaluation on Camera Pose Estimation- Short Sequence} on Sintel \cite{butler2012naturalistic}, TUM-dynamic \cite{sturm2012benchmark}, and ScanNet \cite{dai2017scannet} datasets. TTT3R achieves the best overall performance among online methods, while its accuracy has not yet matched strong offline methods (e.g., VGGT), where full attention, despite being slower and more memory-demanding, preserves the entire history context.}
	\resizebox{\linewidth}{!}{
		\begin{tabular}{lc|ccc|ccc|ccc}
			\toprule
			& & \multicolumn{3}{c|}{\textbf{Sintel (50 frames)}} & \multicolumn{3}{c|}{\textbf{TUM-dynamics (90 frames)}} & \multicolumn{3}{c}{\textbf{ScanNet (90 frames)}} \\
			\textbf{Method} & \textbf{Online} & ATE $\downarrow$ & RPE trans $\downarrow$ & RPE rot $\downarrow$ & ATE $\downarrow$ & RPE trans $\downarrow$ & RPE rot $\downarrow$ & ATE $\downarrow$ & RPE trans $\downarrow$ & RPE rot $\downarrow$ \\
			\midrule
			Robust-CVD \cite{kopf2021robust}  & \xmark & 0.360 & 0.154 & 3.443 & 0.153 & 0.026 & 3.528 & 0.227 & 0.064 & 7.374 \\
			CasualSAM \cite{zhang2022structure}   & \xmark & 0.141 & \textbf{0.035} & \underline{0.615} & 0.071 & \textbf{0.010} & 1.712 & 0.158 & 0.034 & 1.618 \\
			DUSt3R \cite{wang2024dust3r}      & \xmark & 0.417 & 0.250 & 5.796 & 0.083 & 0.017 & 3.567 & 0.081 & 0.028 & 0.784 \\
			MAST3R \cite{leroy2024grounding}      & \xmark & 0.185 & 0.060 & 1.496 & \underline{0.038} & \underline{0.012} & \underline{0.448} & 0.078 & 0.020 & \underline{0.475} \\
			MonST3R \cite{zhang2024monst3r}     & \xmark & \underline{0.111} & 0.044 & 0.869 & 0.098 & 0.019 & 0.935 & 0.077 & 0.018 & 0.529 \\
			Easi3R \cite{chen2025easi3r}      & \xmark & \textbf{0.110} & \underline{0.042} & 0.758 & 0.105 & 0.022 & 1.064 & \underline{0.061} & \underline{0.017} & 0.525 \\
			AETHER \cite{zhu2025aether}       & \xmark & 0.189 & 0.054 & 0.694 & 0.092 & \underline{0.012} & 1.106 & 0.176 & 0.028 & 1.204 \\
			VGGT \cite{wang2025vggt}        & \xmark & 0.172 & 0.062 & \textbf{0.471} & \textbf{0.012} & \textbf{0.010} & \textbf{0.310} & \textbf{0.035} & \textbf{0.015} & \textbf{0.377} \\
			\midrule
			Spann3R \cite{wang20253d}     & \cmark & 0.329 & 0.110 & 4.471 & 0.056 & 0.021 & 0.591 & 0.096 & 0.023 & 0.661 \\
			CUT3R \cite{wang2025continuous}       & \cmark & \underline{0.213} & \underline{0.066} & \underline{0.621} & 0.046 & 0.015 & 0.473 & 0.099 & \underline{0.022} & \underline{0.600} \\
			Point3R \cite{wu2025point3r}     & \cmark & 0.351 & 0.128 & 1.822 & 0.075 & 0.029 & 0.642 & 0.106 & 0.035 & 1.946 \\
			StreamVGGT \cite{zhuo2025streaming}  & \cmark & 0.251 & 0.149 & 1.894 & 0.061 & 0.033 & 3.209 & 0.161 & 0.057 & 3.647 \\
			TTT3R \cite{chen2025ttt3r}  & \cmark & \textbf{0.201} & \textbf{0.063} & \textbf{0.61} & \underline{0.028} & \underline{0.012} & \underline{0.379} & \underline{0.064} & \textbf{0.021} & \textbf{0.592} \\
			\textbf{BAT3R}  & \cmark & 0.219 & 0.089 & 0.84 & \textbf{0.016} & \textbf{0.008} & \textbf{0.287} & \textbf{0.058} & 0.022 & 0.852 \\
			\bottomrule
		\end{tabular}
	}
	\label{short1}
\end{table*}

\begin{figure*}[htbp]
	\centering

	\includegraphics[width=\textwidth]{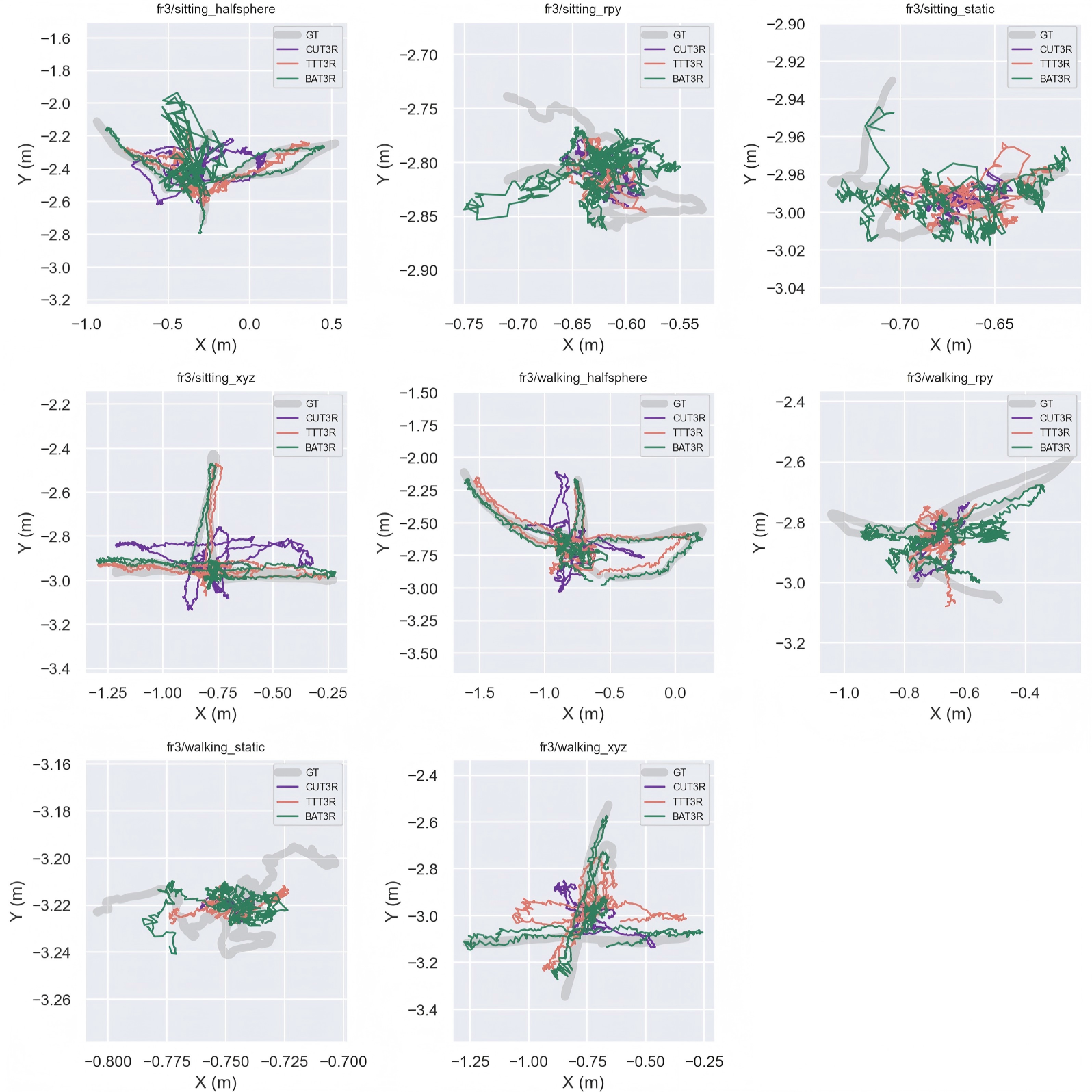}
	\caption{Qualitative comparison of eight estimated camera trajectories on the TUM RGB-D dataset.}
	\label{fig:tum_visualize}
\end{figure*}

\subsection{Camera Pose Estimation}
\label{subsec:pose_eval}
\textbf{Experimental Settings:} Following established protocols \cite{wang2025continuous, zhang2024monst3r}, we evaluate our method on the Sintel, TUM-dynamics, and ScanNet datasets. After applying Sim(3) alignment to the estimated trajectories, we adopt Absolute Translation Error (ATE), Relative Translation Error (RTE), and Relative Rotation Error (RRE) as our evaluation metrics. The compared baselines encompass mainstream 3D reconstruction foundation models, as well as per-sequence optimization approaches, such as RobustCVD \cite{kopf2021robust} and CasualSAM \cite{zhang2022structure}, which jointly optimize camera parameters and dense depth maps.

\textbf{Quantitative Analysis}: The experimental results for the long-sequence evaluation are illustrated in Fig. \ref{fig:pose_dataset}. Since many baseline methods are restricted to short sequences due to out-of-memory (OOM) constraints, we additionally compile the short-sequence comparisons in Table \ref{short1}, explicitly categorizing the baselines into offline methods (which require global attention or extra optimization) and online methods. The quantitative results indicate that our method achieves leading performance among all online approaches on the TUM-dynamics and ScanNet datasets. Specifically, on the TUM-dynamics dataset, BAT3R achieves an Absolute Translation Error (ATE) of 0.016, representing a significant 65.2\% error reduction compared to the CUT3R baseline (0.046) and a 42.9\% improvement over the highly competitive TTT3R (0.028). Furthermore, BAT3R dominates all online methods in rotational and translational precision on this dataset (RPE trans of 0.008 and RPE rot of 0.287). Similarly, on the ScanNet dataset, BAT3R secures the lowest ATE among online methods at 0.058, yielding a 41.4\% reduction relative to CUT3R (0.099) and a 9.4\% reduction compared to TTT3R (0.064). Despite these substantial gains, we observe a remaining performance gap compared to the upper bound established by strong offline methods (e.g., VGGT), which preserve the entire historical context through full attention at the cost of significantly higher memory demands and slower processing speeds.

\textbf{Qualitative Analysis}: Visual comparisons of the camera trajectories in Fig. \ref{fig:tum_visualize} demonstrate that BAT3R yields more accurate and robust pose estimation than TTT3R \cite{chen2025ttt3r} and CUT3R \cite{wang2025continuous}. Specifically, BAT3R (green) consistently maintains high fidelity to the ground truth across various motion patterns, effectively suppressing the cumulative drift observed in the baselines. This performance underscores the efficacy of our BASU module in ensuring global consistency during long-term reconstruction.

\begin{figure*}[htbp]
	\centering
	\includegraphics[width=\textwidth]{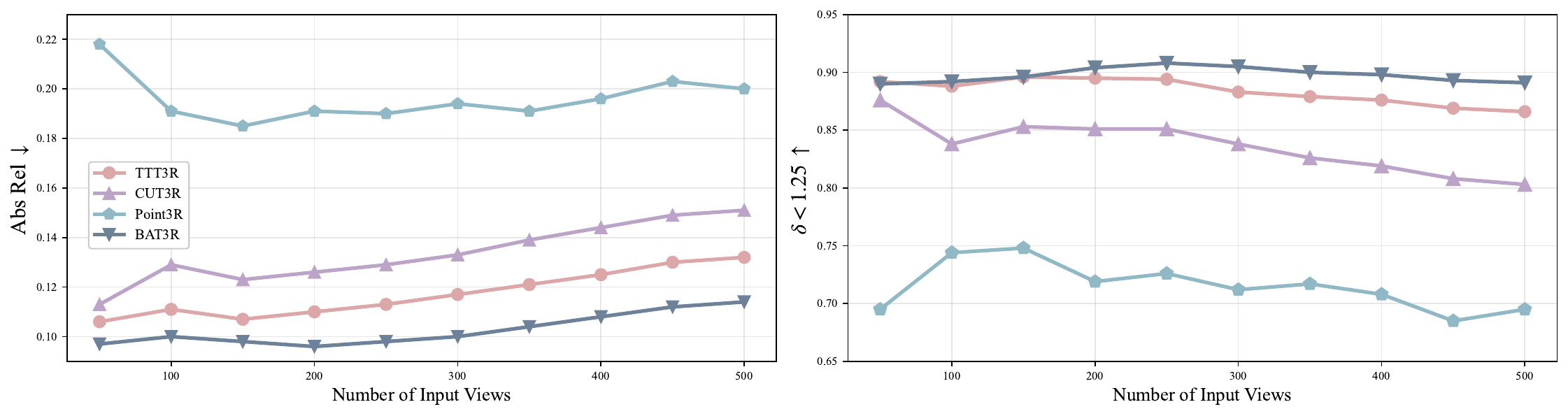}
	\caption{Metric depth evaluation on KITTI, excluding VGGT-based methods that don’t support metric depth.}
	\label{fig:kitti_dataset}
\end{figure*}

\begin{table*}[htbp]
	\centering
	\caption{\textbf{Evaluation of Video Depth Estimation - Short Sequence.} We report scale-invariant relative depth (aligned by a per-sequence scale) and metric scale absolute depth accuracy on Sintel \cite{butler2012naturalistic}, KITTI \cite{geiger2013vision} and Bonn \cite{palazzolo2019refusion} datasets. BAT3R achieves competitive performance among online methods.}
	\resizebox{\linewidth}{!}{
		\begin{tabular}{llc|cc|cc|cc}
			\toprule
			& & & \multicolumn{2}{c|}{\textbf{Sintel (50 frames)}} & \multicolumn{2}{c|}{\textbf{BONN (110 frames)}} & \multicolumn{2}{c}{\textbf{KITTI (110 frames)}} \\
			\cmidrule{4-9}
			\textbf{Alignment} & \textbf{Method} & \textbf{Online} & Abs Rel $\downarrow$ & $\delta<1.25$ $\uparrow$ & Abs Rel $\downarrow$ & $\delta<1.25$ $\uparrow$ & Abs Rel $\downarrow$ & $\delta<1.25$ $\uparrow$ \\
			\midrule
			\multirow{11}{*}{Per-sequence Scale}
			& DUSt3R \cite{wang2024dust3r}     & \xmark & 0.656 & 45.2 & 0.155 & 83.3 & 0.144 & 81.3 \\
			& MAST3R \cite{leroy2024grounding}   & \xmark & 0.641 & 43.9 & 0.252 & 70.1 & 0.183 & 74.5 \\
			& MonST3R \cite{zhang2024monst3r}   & \xmark & 0.378 & 55.8 & 0.067 & 96.3 & 0.168 & 74.4 \\
			& Easi3R \cite{chen2025easi3r}   & \xmark & 0.377 & 55.9 & 0.059 & 97.0 & 0.102 & 91.2 \\
			& AETHER \cite{zhu2025aether}     & \xmark & 0.324 & 50.2 & 0.273 & 59.4 & \textbf{0.056} & \textbf{97.8} \\
			& VGGT \cite{wang2025vggt}      & \xmark & \textbf{0.287} & \textbf{66.1} & \textbf{0.055} & \textbf{97.1} & \underline{0.070} & \underline{96.5} \\
			\cmidrule{2-9}
			& Spann3R \cite{wang20253d}   & \cmark & 0.622 & 42.6 & 0.144 & 81.3 & 0.198 & 73.7 \\
			& CUT3R \cite{wang2025continuous}      & \cmark & 0.421 & 47.9 & 0.078 & 93.7 & 0.118 & 88.1 \\
			& Point3R  \cite{wu2025point3r}   & \cmark & 0.452 & 48.9 & 0.060 & 96.0 & 0.136 & 84.2 \\
			& StreamVGGT \cite{zhuo2025streaming} & \cmark & \textbf{0.323} & \textbf{65.7} & \textbf{0.059} & \textbf{97.2} & 0.173 & 72.1 \\
			& TTT3R \cite{chen2025ttt3r} & \cmark & 0.404 & 50.0 & 0.068 & 95.4 & \underline{0.113} & \underline{90.4} \\
			& \textbf{BAT3R (Ours)} & \cmark & \underline{0.401} & \underline{50.1} & \underline{0.063} & \underline{96.8} & \textbf{0.111} & \textbf{91.1} \\
			\midrule
			\multirow{4}{*}{Metric Scale}
			& MAST3R \cite{leroy2024grounding}    & \xmark & 1.022 & 14.3 & 0.272 & 70.6 & 0.467 & 15.2 \\
			& CUT3R \cite{wang2025continuous}     & \cmark & 1.029 & 23.8 & \underline{0.103} & 88.5 & 0.122 & 85.5 \\
			& Point3R \cite{wu2025point3r}   & \cmark & \textbf{0.777} & 17.1 & 0.137 & \textbf{94.7} & 0.191 & 73.8 \\
			& TTT3R \cite{chen2025ttt3r} & \cmark & 0.977 & \textbf{24.5} & \textbf{0.090} & \underline{94.2} & \underline{0.110} & \underline{89.1} \\
			& \textbf{BAT3R (Ours)} & \cmark & \underline{0.930} & \textbf{24.5} & 0.105 & 88.6 & \textbf{0.099} & \textbf{89.4} \\
			\bottomrule
		\end{tabular}
	}
	\label{short2}
\end{table*}

\subsection{Video Depth Estimation}
\label{subsec:depth_eval}
\textbf{Experimental Settings:} Following standard protocols \cite{wang2025continuous, zhang2024monst3r}, we assess video depth estimation across the KITTI, Sintel, and Bonn datasets, which encompass a diverse mix of dynamic, static, indoor, and outdoor environments. Our primary evaluation metrics are the Absolute Relative Error (Abs Rel) and the $\delta < 1.25$ threshold accuracy. To evaluate relative depth accuracy, capturing both per-frame fidelity and inter-frame consistency, we align the predicted depth maps with the ground truth using a per-sequence scale factor. Additionally, for models capable of generating metric pointmaps (i.e., absolute scale in meters), we report unaligned metrics to rigorously evaluate real-world absolute-scale accuracy.

\textbf{Quantitative Results:} The short-sequence evaluations presented in Table \ref{short2} confirm that our approach consistently maintains state-of-the-art or highly competitive performance among online frameworks entirely without fine-tuning. Specifically, BAT3R secures the leading position among online methods on the KITTI dataset across both metric and scale-invariant settings. Under the metric scale evaluation on KITTI, BAT3R achieves an Abs Rel of 0.099, representing a significant 18.9\% error reduction compared to the CUT3R baseline (0.122) and an additional 10.0\% reduction over the highly competitive TTT3R (0.110). Furthermore, it demonstrates robust generalization on the Sintel and Bonn datasets by frequently ranking in the top two. For instance, in the scale-invariant setting on the Bonn dataset, BAT3R reduces the Abs Rel to 0.063, a 19.2\% relative improvement over CUT3R (0.078). Similarly, under Sintel's metric scale evaluation, BAT3R decreases the absolute relative error by 9.6\% against CUT3R (0.930 vs. 1.029), while simultaneously tying for the highest $\delta<1.25$ threshold accuracy (24.5).

\begin{table*}[t]
	\centering
	\caption{3D Reconstruction Results on 7-Scenes and NRGBD.}
	\label{tab:reconstruction_results}
	\resizebox{\linewidth}{!}{
		\begin{tabular}{ll cccccc cccccc}
			\toprule
			\multirow{3}{*}{Model} & \multirow{3}{*}{Input} & \multicolumn{6}{c}{7-Scenes} & \multicolumn{6}{c}{NRGBD} \\
			\cmidrule(lr){3-8} \cmidrule(lr){9-14}
			& & \multicolumn{2}{c}{Acc. $\downarrow$} & \multicolumn{2}{c}{Comp. $\downarrow$} & \multicolumn{2}{c}{NC $\uparrow$} & \multicolumn{2}{c}{Acc. $\downarrow$} & \multicolumn{2}{c}{Comp. $\downarrow$} & \multicolumn{2}{c}{NC $\uparrow$} \\
			\cmidrule(lr){3-4} \cmidrule(lr){5-6} \cmidrule(lr){7-8} \cmidrule(lr){9-10} \cmidrule(lr){11-12} \cmidrule(lr){13-14}
			& & Mean & Med. & Mean & Med. & Mean & Med. & Mean & Med. & Mean & Med. & Mean & Med. \\
			\midrule
			\multirow{3}{*}{VGGT (Offline) \cite{wang2025vggt}}
			& 300 & \textcolor{gray}{\textit{OOM}} & \textcolor{gray}{\textit{OOM}} & \textcolor{gray}{\textit{OOM}} & \textcolor{gray}{\textit{OOM}} & \textcolor{gray}{\textit{OOM}} & \textcolor{gray}{\textit{OOM}} & \textcolor{gray}{\textit{OOM}} & \textcolor{gray}{\textit{OOM}} & \textcolor{gray}{\textit{OOM}} & \textcolor{gray}{\textit{OOM}} & \textcolor{gray}{\textit{OOM}} & \textcolor{gray}{\textit{OOM}} \\
			& 400 & \textcolor{gray}{\textit{OOM}} & \textcolor{gray}{\textit{OOM}} & \textcolor{gray}{\textit{OOM}} & \textcolor{gray}{\textit{OOM}} & \textcolor{gray}{\textit{OOM}} & \textcolor{gray}{\textit{OOM}} & \textcolor{gray}{\textit{OOM}} & \textcolor{gray}{\textit{OOM}} & \textcolor{gray}{\textit{OOM}} & \textcolor{gray}{\textit{OOM}} & \textcolor{gray}{\textit{OOM}} & \textcolor{gray}{\textit{OOM}} \\
			& 500 & \textcolor{gray}{\textit{OOM}} & \textcolor{gray}{\textit{OOM}} & \textcolor{gray}{\textit{OOM}} & \textcolor{gray}{\textit{OOM}} & \textcolor{gray}{\textit{OOM}} & \textcolor{gray}{\textit{OOM}} & \textcolor{gray}{\textit{OOM}} & \textcolor{gray}{\textit{OOM}} & \textcolor{gray}{\textit{OOM}} & \textcolor{gray}{\textit{OOM}} & \textcolor{gray}{\textit{OOM}} & \textcolor{gray}{\textit{OOM}} \\
			\midrule
			\multirow{3}{*}{StreamVGGT \cite{zhuo2025streaming}}
			& 300 & \textcolor{gray}{\textit{OOM}} & \textcolor{gray}{\textit{OOM}} & \textcolor{gray}{\textit{OOM}} & \textcolor{gray}{\textit{OOM}} & \textcolor{gray}{\textit{OOM}} & \textcolor{gray}{\textit{OOM}} & \textcolor{gray}{\textit{OOM}} & \textcolor{gray}{\textit{OOM}} & \textcolor{gray}{\textit{OOM}} & \textcolor{gray}{\textit{OOM}} & \textcolor{gray}{\textit{OOM}} & \textcolor{gray}{\textit{OOM}} \\
			& 400 & \textcolor{gray}{\textit{OOM}} & \textcolor{gray}{\textit{OOM}} & \textcolor{gray}{\textit{OOM}} & \textcolor{gray}{\textit{OOM}} & \textcolor{gray}{\textit{OOM}} & \textcolor{gray}{\textit{OOM}} & \textcolor{gray}{\textit{OOM}} & \textcolor{gray}{\textit{OOM}} & \textcolor{gray}{\textit{OOM}} & \textcolor{gray}{\textit{OOM}} & \textcolor{gray}{\textit{OOM}} & \textcolor{gray}{\textit{OOM}} \\
			& 500 & \textcolor{gray}{\textit{OOM}} & \textcolor{gray}{\textit{OOM}} & \textcolor{gray}{\textit{OOM}} & \textcolor{gray}{\textit{OOM}} & \textcolor{gray}{\textit{OOM}} & \textcolor{gray}{\textit{OOM}} & \textcolor{gray}{\textit{OOM}} & \textcolor{gray}{\textit{OOM}} & \textcolor{gray}{\textit{OOM}} & \textcolor{gray}{\textit{OOM}} & \textcolor{gray}{\textit{OOM}} & \textcolor{gray}{\textit{OOM}} \\
			\midrule
			\multirow{3}{*}{CUT3R \cite{wang2025continuous}}
			& 300 & 0.141 & 0.096 & 0.076 & 0.034 & 0.543 & 0.564 & 0.234 & 0.139 & 0.074 & 0.018 & 0.575 & 0.614 \\
			\cmidrule{2-14}
			& 400 & 0.178 & 0.121 & 0.115 & 0.069 & 0.532 & 0.546 & 0.342 & 0.227 & 0.127 & 0.067 & 0.561 & 0.591 \\
			\cmidrule{2-14}
			& 500 & 0.190 & 0.138 & 0.090 & 0.033 & 0.530 & 0.543 & 0.359 & 0.264 & 0.173 & 0.081 & 0.560 & 0.591 \\
			\midrule
			\multirow{3}{*}{TTT3R \cite{chen2025ttt3r}}
			& 300 & 0.040 & 0.025 & 0.024 & 0.005 & 0.567 & 0.602 & 0.101 & 0.044 & 0.025 & 0.005 & 0.610 & 0.678 \\
			\cmidrule{2-14}
			& 400 & 0.052 & 0.031 & 0.027 & 0.005 & 0.558 & 0.588 & 0.143 & 0.065 & 0.071 & 0.012 & 0.600 & 0.658 \\
			\cmidrule{2-14}
			& 500 & 0.066 & 0.039 & 0.031 & 0.006 & 0.551 & 0.577 & 0.166 & 0.092 & 0.087 & 0.021 & 0.593 & 0.647 \\
			\midrule
			\multirow{3}{*}{TTSA3R \cite{zheng2026ttsa3r}}
			& 300 & 0.036 & 0.020 & 0.035 & 0.006 & 0.566 & 0.600 & 0.090 & 0.036 & 0.020 & 0.004 & 0.620 & 0.696 \\
			\cmidrule{2-14}
			& 400 & 0.036 & 0.019 & 0.024 & 0.004 & 0.561 & 0.592 & 0.104 & 0.045 & 0.035 & 0.006 & 0.618 & 0.692 \\
			\cmidrule{2-14}
			& 500 & 0.042 & 0.021 & 0.024 & 0.004 & 0.556 & 0.585 & 0.121 & 0.054 & 0.050 & 0.006 & 0.613 & 0.684 \\
			\midrule
			\multirow{3}{*}{BAT3R}
			& 300 & \textbf{0.021} & \textbf{0.010} & \textbf{0.021} & \textbf{0.004} & \textbf{0.567} & \textbf{0.602} & \textbf{0.082} & \textbf{0.028} & \textbf{0.014} & \textbf{0.004} & \textbf{0.625} & \textbf{0.705} \\
			\cmidrule{2-14}
			& 400 & \textbf{0.025} & \textbf{0.012} &\textbf{ 0.023} & \textbf{0.004} & \textbf{0.561} & \textbf{0.592} & \textbf{0.093} & \textbf{0.037} & \textbf{0.027} & \textbf{0.004} & \textbf{0.620} & \textbf{0.697} \\
			\cmidrule{2-14}
			& 500 & \textbf{0.026} & \textbf{0.012} & \textbf{0.023} & \textbf{0.004} & \textbf{0.558} & \textbf{0.588} & \textbf{0.097} & \textbf{0.037} & \textbf{0.032} & \textbf{0.004} & \textbf{0.615} & \textbf{0.687} \\
			\bottomrule
		\end{tabular}
	}
\end{table*}

\subsection{3D Reconstruction}
\label{subsec:recon_eval}
To provide a comprehensive quantitative analysis, we evaluate the models across multiple geometric dimensions, including Accuracy, Completeness, and Normal Consistency on the 7-Scenes \cite{shotton2013scene} and NRGBD \cite{azinovic2022neural} benchmarks. As demonstrated in the results, BAT3R consistently outperforms existing recurrent baselines across all metrics and sequence lengths. On the 7-Scenes dataset at 500 frames, BAT3R achieves a Mean Accuracy error of 0.026, which represents an approximate 86.3\% error reduction compared to the CUT3R baseline (0.190) and a 60.6\% reduction compared to the highly competitive TTT3R (0.066). Furthermore, BAT3R demonstrates exceptional performance on the NRGBD dataset at 500 frames, achieving a Mean Accuracy of 0.097 and a Mean Completeness of 0.032. This corresponds to a 41.6\% decrease in accuracy error and a 63.2\% decrease in completeness error relative to TTT3R. Even when compared to the recent TTSA3R \cite{zheng2026ttsa3r} method, BAT3R yields robust error reductions across accuracy and completeness while simultaneously securing the highest Normal Consistency scores (e.g., 0.615 on NRGBD at 500 frames) across both datasets.

\begin{figure*}[t]
	\centering

	\begin{subfigure}[b]{0.32\textwidth}
		\centering
		\includegraphics[width=\textwidth]{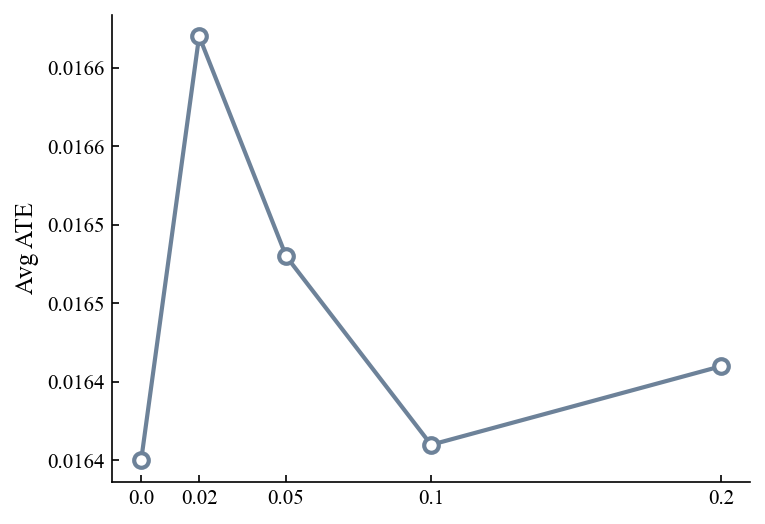}
		\caption{}
		\label{fig:abl_floor}
	\end{subfigure}
	\hfill

	\begin{subfigure}[b]{0.32\textwidth}
		\centering
		\includegraphics[width=\textwidth]{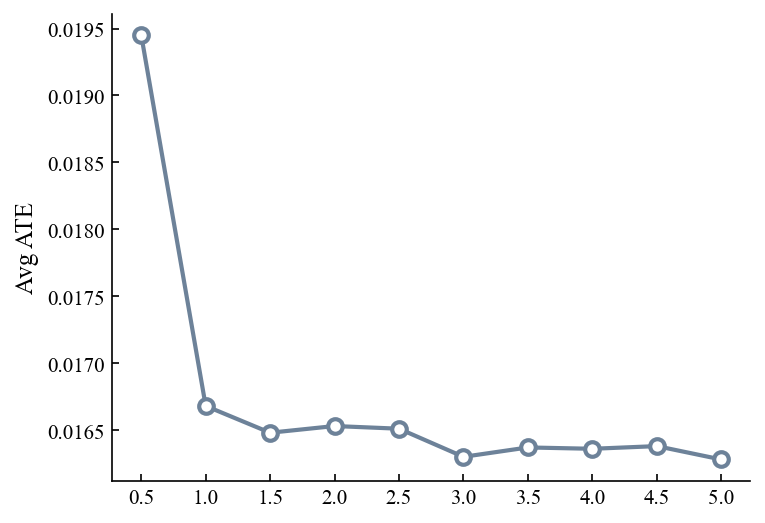}
		\caption{}
		\label{fig:abl_slope}
	\end{subfigure}
	\hfill

	\begin{subfigure}[b]{0.32\textwidth}
		\centering
		\includegraphics[width=\textwidth]{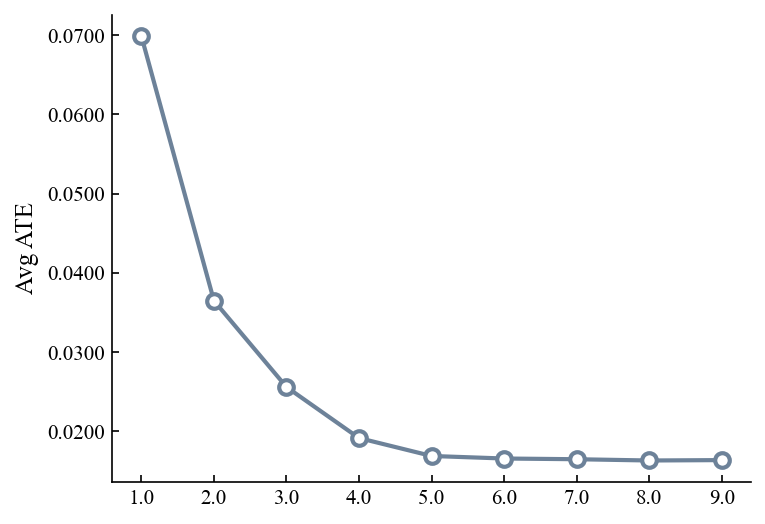}
		\caption{}
		\label{fig:abl_threshold}
	\end{subfigure}

	\caption{\textbf{Hyperparameter sensitivity analysis} defined in Eq.~\ref{eq:sig_gate}. We report the Average ATE to evaluate the impact of (a) the minimum update floor $\eta$, (b) the scaling factor $\kappa$ controlling the transition slope, and (c) the shifting threshold $\tau$. }
	\label{fig:overall_ablation}
\end{figure*}

\begin{figure*}[t]
	\centering

	\begin{subfigure}[b]{0.32\textwidth}
		\centering
		\includegraphics[width=\textwidth]{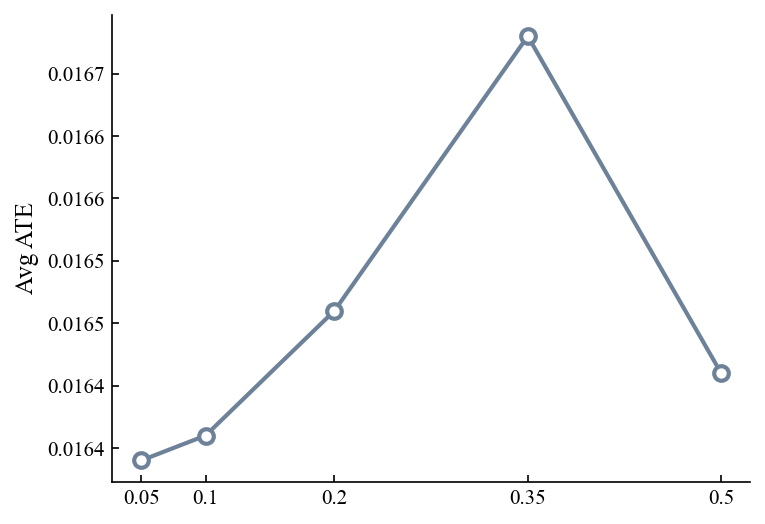}
		\caption{Impact of rel\_floor parameter}
		\label{fig:abl_rel_floor}
	\end{subfigure}
	\hfill

	\begin{subfigure}[b]{0.33\textwidth}
		\centering
		\includegraphics[width=\textwidth]{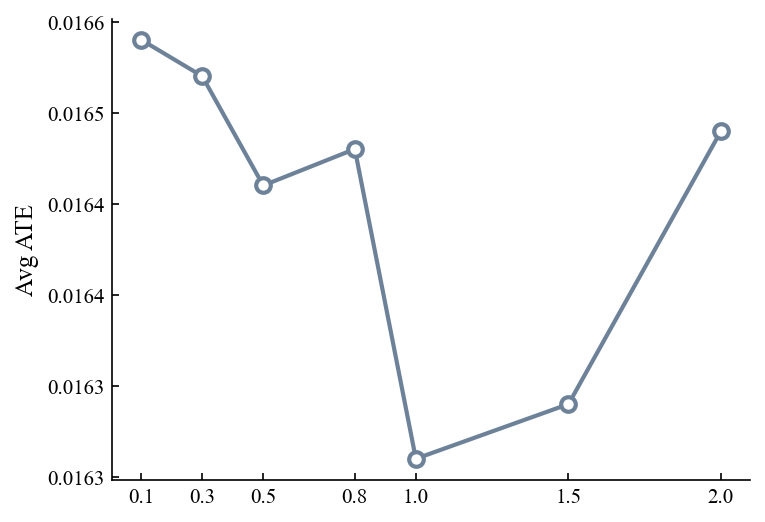}
		\caption{Impact of APN alpha parameter}
		\label{fig:abl_apn_alpha}
	\end{subfigure}
	\hfill

	\begin{subfigure}[b]{0.33\textwidth}
		\centering
		\includegraphics[width=\textwidth]{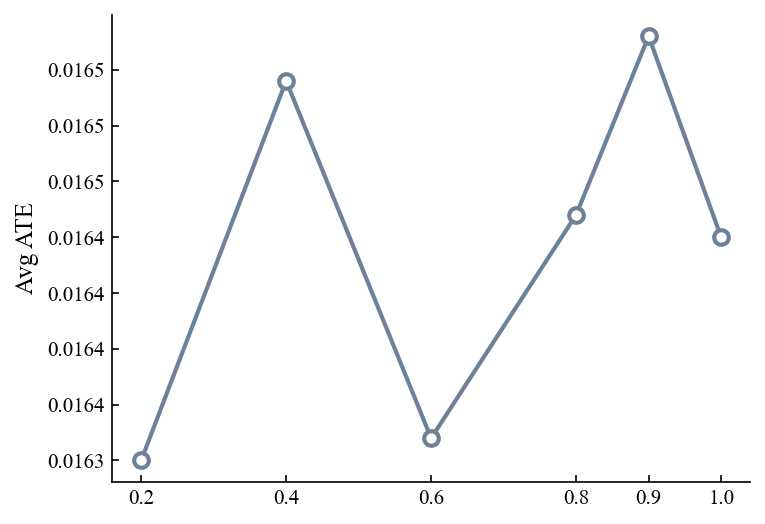}
		\caption{Impact of KG weight parameter}
		\label{fig:abl_kg_weight}
	\end{subfigure}

	\caption{Ablation studies on additional hyperparameters. We report the Average ATE for (a) relative floor, (b) APN alpha, and (c) KG weight settings to evaluate their sensitivity.}
	\label{fig:additional_ablation}
\end{figure*}

\subsection{Parameter Analysis}
\label{subsec:param_analysis}

To comprehensively evaluate the robustness of the BAT3R framework, we conduct detailed hyperparameter sensitivity experiments on the TUM-dynamics dataset, utilizing Average ATE as the primary evaluation metric. We systematically analyze the critical parameters within both the State Innovation Gating (SIG) and Bayesian-guided Adaptive State Updating (BASU) modules.

\textbf{SIG Hyperparameters:} We assess the impact of the minimum update floor ($\eta$), the scaling factor ($\kappa$), and the shifting threshold ($\tau$) defined in Eq.~\ref{eq:sig_gate}. As illustrated in Fig. \ref{fig:overall_ablation}(a), the Average ATE is relatively insensitive to the update floor $\eta$, maintaining highly stable performance with only minor fluctuations across the evaluated range. In contrast, the scaling factor $\kappa$ (Fig. \ref{fig:overall_ablation}(b)) and the threshold $\tau$ (Fig. \ref{fig:overall_ablation}(c)) both exhibit a sharp initial performance gain. Specifically, the ATE drops significantly and then plateaus once $\kappa$ exceeds $1.0$ and $\tau$ exceeds $5.0$. This indicates that the gating mechanism requires a sufficiently steep slope and a wide margin to effectively bound representational drift.

\textbf{BASU Hyperparameters:} We further evaluate the additional hyperparameters within the BASU module, including the relative covariance floor, the Adaptive Process Noise (APN) scaling parameter ($\alpha$), and the Kalman Gain (KG) interpolation weight ($\omega$), as shown in Fig. \ref{fig:additional_ablation}. First, Fig. \ref{fig:additional_ablation}(a) demonstrates the impact of the minimum threshold bounding the state covariance; the model performs optimally at an extremely low value ($0.0$), allowing the covariance to shrink organically as confident observations accumulate, whereas setting it marginally higher (e.g., $0.02$) artificially inflates the lower bound and causes a temporary spike in ATE before stabilizing near $0.1$. Second, the APN $\alpha$ parameter dictates how aggressively process noise is inflated in response to inter-frame motion. As depicted in Fig. \ref{fig:additional_ablation}(b), the model achieves optimal performance precisely at $1.0$, since lower values fail to accommodate rapid motion, while higher values inject excessive uncertainty into the latent state. Finally, Fig. \ref{fig:additional_ablation}(c) evaluates the KG interpolation weight ($\omega$) between the standard ratio and square-root Kalman gain formulations. The performance shows minor fluctuations, yielding the best results at $0.2$ and $0.6$, which suggests that blending the two formulations—leaning slightly towards the square-root variant—provides optimal numerical stability when observation noise and state covariance differ significantly.

\section{Theoretical Analysis}
\label{sec:theoretical_analysis}

In this section, we provide the formal theoretical foundation for the BAT3R framework, specifically focusing on the stability and optimality of our Bayesian-guided state transition rules.

\subsection{Stability Analysis of SIG}
\label{subsec:theory_sig}
The SIG module acts as a distributional regularizer by penalizing states that deviate excessively from the initial geometric embedding $\mathbf{S}_0$. We prove that this mechanism ensures the latent state remains within a bounded region of the decoder's reliable latent manifold.

\textbf{Theorem 1 (Contraction toward Anchor).} \textit{Let $\delta_t = ||\mathbf{S}_t' - \mathbf{S}_0||_2$ be the innovation residual at time $t$. Under the SIG update rule $\widetilde{\mathbf{S}}_t = g_t \mathbf{S}_t' + (1-g_t) \mathbf{S}_{t-1}$, where $g_t$ is a decreasing function of $\delta_t$, the SIG module acts as a non-expansive mapping that constrains the state trajectory relative to the initial anchor $\mathbf{S}_0$.}

\textit{Proof.} Consider the distance to the anchor after the SIG update:
\begin{equation}
	||\widetilde{\mathbf{S}}_t - \mathbf{S}_0||_2 = ||g_t \mathbf{S}_t' + (1-g_t) \mathbf{S}_{t-1} - \mathbf{S}_0||_2
\end{equation}
By the triangle inequality:
\begin{equation}
	||\widetilde{\mathbf{S}}_t - \mathbf{S}_0||_2 \leq g_t ||\mathbf{S}_t' - \mathbf{S}_0||_2 + (1-g_t) ||\mathbf{S}_{t-1} - \mathbf{S}_0||_2
\end{equation}
Substituting the innovation $\delta_t$:
\begin{equation}
	||\widetilde{\mathbf{S}}_t - \mathbf{S}_0||_2 \leq g_t \delta_t + (1-g_t) ||\mathbf{S}_{t-1} - \mathbf{S}_0||_2
\end{equation}
Since the sigmoid gate $g_t \to \eta$ (lower bound) as $\delta_t \to \infty$, the update weight shifts predominantly toward the previous state $\mathbf{S}_{t-1}$ whenever the candidate $\mathbf{S}_t'$ exhibits anomalous drift. This ensures that the state evolution is not solely driven by potentially noisy current observations but is regularized by the historical trajectory and the geometric anchor $\mathbf{S}_0$. Thus, the latent representation is prevented from migrating into regions where the frozen decoder lacks reconstruction fidelity.

\subsection{Geometric Manifold Consistency}
\label{subsec:theory_manifold}
A fundamental limitation of previous recurrent architectures like CUT3R and TTT3R is the unbounded accumulation of error variance over long sequences, leading to structural distortion. We formally prove that the BASU module strictly bounds the state uncertainty.

\textbf{Theorem 2 (Uncertainty Boundedness).} \textit{Consider the recurrent state update in BASU where the process noise covariance $\mathbf{Q}_t$ is bounded by $||\mathbf{Q}_t||_2 \leq Q_{max}$ and the observation noise covariance $\mathbf{R}_t$ (derived from the confidence map $C_t$) is bounded by $||\mathbf{R}_t||_2 \leq R_{max}$. For any sequence length $t > 0$, the spectral norm of the posterior state covariance $\boldsymbol{\Sigma}_{t|t}$ is strictly upper-bounded by the maximum observation noise.}

\textit{Proof.} From the recursive covariance update, the prior covariance is $\boldsymbol{\Sigma}_{t|t-1} = \boldsymbol{\Sigma}_{t-1|t-1} + \mathbf{Q}_t$. The posterior covariance update can be reformulated using the matrix inversion lemma as:
\begin{equation}
	\boldsymbol{\Sigma}_{t|t} = (\boldsymbol{\Sigma}_{t|t-1}^{-1} + \mathbf{R}_t^{-1})^{-1}
\end{equation}
Given that $\boldsymbol{\Sigma}_{t|t-1}$ is a positive-definite matrix resulting from the accumulation of initial variance $\boldsymbol{\Sigma}_{0|0}$ and process noise $\mathbf{Q}_t$, its inverse $\boldsymbol{\Sigma}_{t|t-1}^{-1}$ is also positive-definite with $||\boldsymbol{\Sigma}_{t|t-1}^{-1}||_2 > 0$. Applying the properties of the spectral norm for positive-definite matrices:
\begin{equation}
	||\boldsymbol{\Sigma}_{t|t}||_2 = \left|\left| \left( \boldsymbol{\Sigma}_{t|t-1}^{-1} + \mathbf{R}_t^{-1} \right)^{-1} \right|\right|_2 < ||(\mathbf{R}_t^{-1})^{-1}||_2 = ||\mathbf{R}_t||_2 \leq R_{max}
\end{equation}
This inequality holds for all $t$ as long as incoming observations are constrained by the confidence-driven noise model $\mathbf{R}_t$. Thus, the representational uncertainty of BAT3R does not grow with sequence length $t$, preventing the catastrophic drift observed in heuristic gating methods.

\subsection{Bayesian Fusion Optimality}
\label{subsec:theory_bayesian}
We further justify that the BASU update rule provides the optimal integration of novel scene geometry.

\textbf{Theorem 3 (Minimum MSE Optimality).} \textit{Under the assumption that the latent state innovation follows a Gaussian distribution, the Kalman gain $\mathbf{K}_t$ derived in BASU yields the Best Linear Unbiased Estimator (BLUE) for the scene state $\mathbf{S}_t^*$, minimizing the expected Mean Squared Error (MSE) $\mathbb{E}[||\mathbf{S}_t^* - \boldsymbol{\mu}_{t|t}||_2^2]$.}

\textit{Proof.} Let the error of the posterior estimate at time $t$ be $\mathbf{e}_t = \mathbf{S}_t^* - \boldsymbol{\mu}_{t|t}$. The objective is to minimize the trace of the posterior covariance matrix $J = \text{Tr}(\boldsymbol{\Sigma}_{t|t})$. Substituting the update rule $\boldsymbol{\mu}_{t|t} = \boldsymbol{\mu}_{t-1|t-1} + \mathbf{K}_t(\widetilde{\mathbf{S}}_t - \boldsymbol{\mu}_{t-1|t-1})$ into the error term, we obtain the posterior covariance $\boldsymbol{\Sigma}_{t|t}$ as a function of $\mathbf{K}_t$:
\begin{equation}
	\boldsymbol{\Sigma}_{t|t} = (\mathbf{I} - \mathbf{K}_t)\boldsymbol{\Sigma}_{t|t-1}(\mathbf{I} - \mathbf{K}_t)^\top + \mathbf{K}_t\mathbf{R}_t\mathbf{K}_t^\top
\end{equation}
Taking the derivative of the trace $\frac{\partial \text{Tr}(\boldsymbol{\Sigma}_{t|t})}{\partial \mathbf{K}_t}$ and setting it to zero:
\begin{equation}
	-2(\mathbf{I} - \mathbf{K}_t)\boldsymbol{\Sigma}_{t|t-1} + 2\mathbf{K}_t\mathbf{R}_t = 0
\end{equation}
Solving for $\mathbf{K}_t$ yields the optimal gain $\mathbf{K}_t = \boldsymbol{\Sigma}_{t|t-1}(\boldsymbol{\Sigma}_{t|t-1} + \mathbf{R}_t)^{-1}$. While BAT3R utilizes a square-root stabilized variant to handle the scale mismatches in high-dimensional transformer latent spaces, it remains anchored to this MSE-minimizing objective, ensuring that well-consolidated memory is protected while novel content is optimally assimilated.

\subsection{Information-Theoretic Gain and Update}
\label{subsec:theory_info}
BASU utilizes a temporal reliability gate $\lambda_t$ derived from the symmetric KL divergence (Information Gain, $IG_t$) between the prior and posterior distributions. We prove that this formulation yields a robust filter that naturally suppresses both redundant observations and anomalous outliers.

\textbf{Theorem 4 (Update Informativeness).} \textit{The information gain $IG_t = D_{KL}(q_{t|t} || q_{t|t-1}) + D_{KL}(q_{t|t-1} || q_{t|t})$ is minimized when the current observation $\widetilde{\mathbf{S}}_t$ provides no novel information relative to the existing state uncertainty, and maximized when the observation signals a significant but reliable structural change.}

\textit{Proof.} For Gaussian beliefs, the symmetric KL divergence is given by Eq. 13 in the main text. Consider two extreme cases:
\begin{itemize}
	\item \textbf{Redundancy}: If $\widetilde{\mathbf{S}}_t \approx \boldsymbol{\mu}_{t-1|t-1}$ and $\mathbf{R}_t \gg \boldsymbol{\Sigma}_{t|t-1}$, the posterior $\boldsymbol{\mu}_{t|t} \to \boldsymbol{\mu}_{t-1|t-1}$ and $\boldsymbol{\Sigma}_{t|t} \to \boldsymbol{\Sigma}_{t|t-1}$. In this limit, $IG_t \to 0$, leading to a reliability gate $\lambda_t \to 0$ via the sigmoid squashing function. This prevents redundant computations and preserves well-consolidated memory.
	\item \textbf{Anomaly}: If $||\widetilde{\mathbf{S}}_t - \boldsymbol{\mu}_{t-1|t-1}||_2 \to \infty$ while $\mathbf{R}_t$ is small, $IG_t$ grows quadratically with the innovation residual. By mapping $IG_t$ through a sigmoid with a high-gain threshold, BAT3R penalizes these anomalously large updates as potential outliers (e.g., dynamic occlusions), effectively shielding the persistent state from corruption.
\end{itemize}
The dual-suppression mechanism ensures that the state plasticity is selectively preserved only for genuinely informative updates that reflect consistent scene geometry.

\end{document}